%% file: main.tex
\documentclass[10pt,twocolumn,letterpaper]{article}

\usepackage{cvpr}              % To produce the CAMERA-READY version
\input{preamble}
\definecolor{cvprblue}{rgb}{0.21,0.49,0.74}
\usepackage[pagebackref,breaklinks,colorlinks,allcolors=cvprblue]{hyperref}

\usepackage{graphicx}
\usepackage[table]{xcolor}   % 为表格启用颜色
\usepackage{colortbl}        % \cellcolor/\columncolor
\usepackage[normalem]{ulem}

\usepackage{wrapfig}
\usepackage{algorithm}   % 提供浮动体环境 \begin{algorithm}
\usepackage{algorithmic} % 提供 \REQUIRE \ENSURE \STATE \FOR \ENDFOR 等

\usepackage{array}           % 自定义列类型 >{...}
\usepackage{booktabs}        % \toprule \midrule \bottomrule
\usepackage{multirow}        % \multirow
\usepackage{makecell}        % \makecell 等
\usepackage{amssymb}  % 提供 \checkmark
\usepackage{bbding}
\usepackage{pifont}   % 提供 \ding{51}/\ding{55} 等
\usepackage{enumitem}  % 控制列表间距（可选）
\usepackage{subcaption}
\usepackage{amsmath}
\DeclareCaptionLabelFormat{custom}{\tablename~\thetable(#2)} % 定义新的caption格式

\usepackage{hyperref}
\usepackage{url}
\usepackage{svrsymbols}
\definecolor{myBrownYellow}{RGB}{204,153,0}
\definecolor{aliceblue}{RGB}{230,255,255}

\newcommand{\CC}{\cellcolor{aliceblue}}

\definecolor{C3}{rgb}{0.839216,0.152941,0.156863}
\newcommand\fixxie[1]{\iffalse [Xie: #1] \fi}
\newcommand\revise[1]{{\color{black}#1}}
\newcommand\Rui[1]{{\color{black}#1}}
\newcommand\RuiR[1]{{\color{black}#1}}

\newcolumntype{a}{>{\columncolor{aliceblue}}r} % 或换成你需要的对齐 l/c/r

\def\paperID{36500} % *** Enter the Paper ID here
\def\confName{CVPR}
\def\confYear{2026}

\title{Accelerating Diffusion Model Training under Minimal Budgets: \\ A Condensation-Based Perspective}

\author{
\textbf{Rui Huang}$^{1, 2\star}$ \quad
\textbf{Shitong Shao}$^{1\star}$ \quad
\textbf{Zikai Zhou}$^{1}$ \quad
\textbf{Pukun Zhao}$^{1}$ \quad
\textbf{Hangyu Guo}$^{3}$ \\
\textbf{Tian Ye}$^{1}$ \quad
\textbf{Lichen Bai}$^{1}$ \quad
\textbf{Shuo Yang}$^{3}$ \quad
\textbf{Zeke Xie}$^{1\dagger}$ \\
$^1$ xLeaF Lab, HKUST (GZ),
$^2$ UESTC, 
$^3$ HIT (SZ) \\
}
\begin{document}
\maketitle
\begingroup
\renewcommand\thefootnote{}\footnotetext{
$^\star$ Equal contribution. \\
\quad \quad $^\dagger$ Correspondence to \textit{zekexie@hkust-gz.edu.cn}. 
}
\begin{figure*}[htb]
  \centering\includegraphics[width=0.905\linewidth]{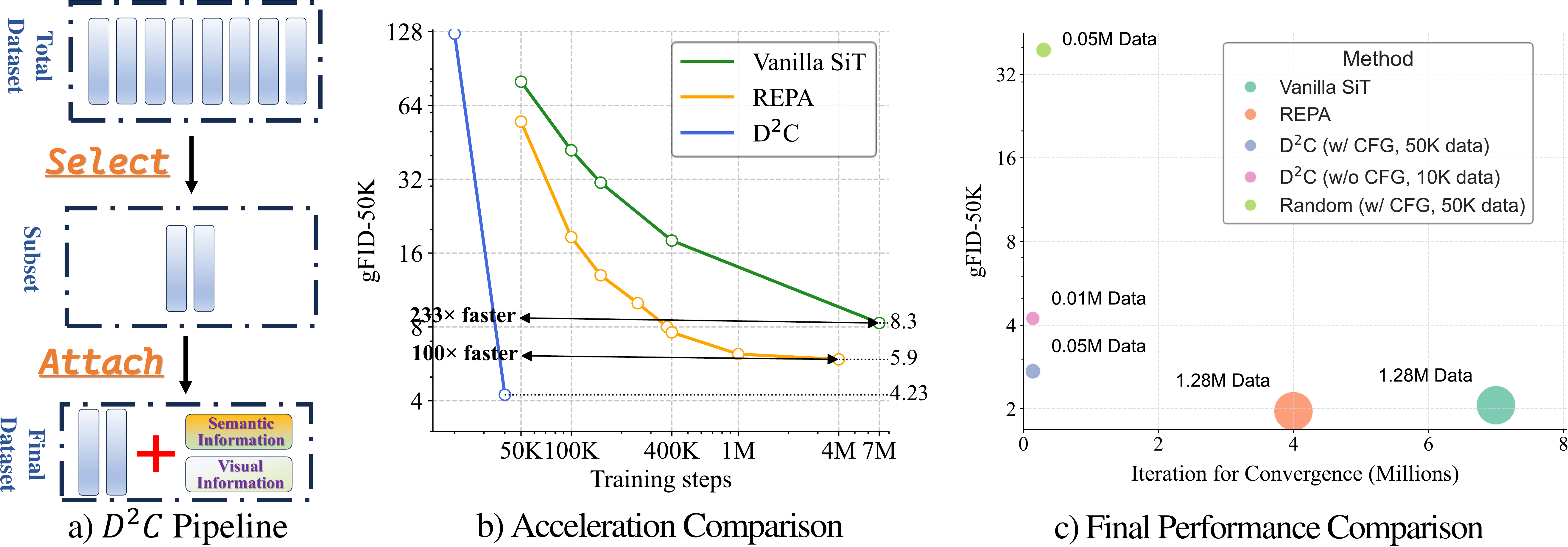}
  \vspace{-6pt}
    \caption{ \textbf{\textit{D$^2$C} framework significantly accelerates diffusion model training with limited data.} \textbf{(a)} Overview of our \textit{D$^2$C} pipeline, which consists of a \textit{Select} phase that filters a compact and diverse subset via diffusion difficulty score and interval sampling, and an \textit{Attach} phase that enriches samples with semantic and visual information. \textbf{(b)} \textit{D$^2$C} achieves over 100$\times$ faster convergence compared to REPA and over 233$\times$ faster than vanilla SiT-XL/2, reaching a FID of 4.3 at just 40k steps. \textbf{(c)} Under a strict 4\% data budget (0.05M), our method achieves a FID of 2.7 at 180k iterations, demonstrating its strong training efficiency and rapid convergence.
    \label{fig:first}
}
\vspace{-15pt}
\end{figure*}

\RuiR{\input{sec/0_abstract}}

\input{sec/1_intro}

\input{sec/2_Preliminaries}

\input{sec/3_method}

\input{sec/4_experiment}
\input{sec/5_conclusion}
{
    \small
    \bibliographystyle{ieeetr}
    \bibliography{main}
}

% WARNING: do not forget to delete the supplementary pages from your submission 

\input{sec/X_suppl}

\end{document}

%% file: sec/0_abstract.tex
\begin{abstract}
Diffusion models have achieved remarkable performance on a wide range of generative tasks, yet training them from scratch is notoriously resource-intensive, typically requiring millions of training images and many GPU days.
Motivated by a data-centric view of this bottleneck, we adopt a condensation-based perspective: given a large training set, the goal is to construct a much smaller \emph{condensed dataset} that still supports training strong diffusion models under minimal data and compute budgets. To operationalize this perspective, we introduce \textbf{D}iffusion \textbf{D}ataset \textbf{C}ondensation (\textbf{\textit{D\textsuperscript{2}C}}), a two-phase framework comprising \textit{Select} and \textit{Attach}. In the \textit{Select} phase, a diffusion difficulty score combined with interval sampling is used to identify a compact, informative training subset from the original data. Building on this subset, the \textit{Attach} phase further strengthens the conditional signals by augmenting each selected image with rich semantic and visual representations. To our knowledge, \textit{D\textsuperscript{2}C} is the first framework that systematically investigates dataset condensation for diffusion models, whereas prior condensation methods have mainly targeted discriminative architectures. Extensive experiments across data budgets (0.8\%–8\% of ImageNet), model architectures, and image resolutions demonstrate that \textbf{\textit{D\textsuperscript{2}C}} dramatically accelerates diffusion model training while preserving high generative quality. On ImageNet 256$^2$ with SiT-XL/2, \textit{D\textsuperscript{2}C} attains a FID of 4.3 in just 40k steps using only 0.8\% of the training images, corresponding to about 233$\times$ and 100$\times$ faster training than vanilla SiT-XL/2 and SiT-XL/2 + REPA, respectively.
\end{abstract}

%% file: sec/1_intro.tex
\section{Introduction}
\label{sec:intro}

Generative models, such as \revise{score-based}~\citep{sde,ddim,ddpm_begin} and flow-based~\citep{iclr22_rect} approaches, have achieved remarkable success in various generative tasks~\citep{chen2025taming}, producing high-quality and diverse \revise{data} across domains~\citep{karras2022elucidating,animatediff,chen2025s2guidancestochasticselfguidance}. However, these \revise{approaches} are notoriously data and compute intensive to train, often requiring millions of samples and hundreds of thousands of iterations to capture complex high-dimensional distributions~\citep{DIT,REPA,SiT}. \revise{The resulting cost presents a significant barrier to broader application and iteration within the AIGC community, making efficient training increasingly important across both academic and industrial settings~\citep{shitong2026fastlightgen,liu2025freqcaacceleratingdiffusionmodels, gu2026mano}.} Recent efforts have improved diffusion training efficiency through \revise{various strategies, such as} architectural redesigns~\citep{SiT,zheng2023fast,DIT}, attention optimization~\citep{tomesd,li2026pisa}, reweighting strategies~\citep{hang2023efficient}, and representation learning~\citep{REPA,REG, CRAFT}. In parallel, data-centric approaches such as patch-based methods~\citep{ding2023patched,wang2023patch}, Infobatch~\citep{qin2023infobatch} and Reweighting\cite{li2025pruning} aim to better \revise{exploit the potential of existing data}. \RuiR{Despite these advances, directly constructing a much smaller yet informative training subset as a primary \emph{condensation-based} route to accelerate diffusion training remains largely unexplored, even though it provides a particularly direct and effective way to reduce data budgets.}

 % \revise{Despite these advances, building a relatively complete ``synthetic\footnote{{Throughout this paper, ``synthetic'' subset refers to an \emph{artificially designed-and-enhanced} subset: real samples are \textbf{\textit{selected}} and \textbf{\textit{augmented/attached}} with semantic and visual representations.}}'' subset via dataset condensation~\citep{dd_begin} remains underexplored.}

% \RuiR{while offering a direct, effective, scalable, widely applicable, and more fundamental data-side route to accelerate diffusion training.}

\RuiR{Dataset condensation~\citep{wu2025dc3,dd_CAFE,dd_sre2l,shao2023generalized} aims to construct a much smaller \emph{condensed} sub-dataset with significantly fewer samples than the original dataset, such that a model trained from scratch on this subset achieves performance comparable to one trained on the full dataset while converging much faster. In practice, existing DC methods typically instantiate this objective in two ways~\citep{wu2025dc3}: (i) \emph{pixel-level} dataset distillation, which directly optimizes synthetic images~\citep{dd_sre2l,dd_mtt}, and (ii) \emph{image-level} condensation, which operates on real images through selection and transformation~\citep{RDED,gao2025principled,OD3}. Unlike classical data pruning or selection~\citep{datasetpruning}, which passively select a fixed subset of existing samples, both branches of dataset condensation actively construct condensed training sets, either by optimizing synthetic images or by selecting and enriching real ones, thereby enabling more aggressive data reduction and higher training efficiency~\citep{RDED}. However, these methods have been developed almost exclusively for discriminative tasks. Compared to discriminative learning, generative diffusion training is substantially more complex and demands higher dataset quality~\citep{manduchi2024challenges}; directly applying popular DC algorithms (e.g., SRe$^2$L~\citep{dd_sre2l}, RDED~\citep{RDED}) to diffusion models often leads to synthetic images that lack structural and semantic fidelity, resulting in degraded sample quality and unstable convergence (see Sec.~\ref{sec:experiment}).}

We raise a key question: \textit{``Can we train diffusion models dramatically faster with significantly less data, while retaining high generation quality?''} The answer is affirmative. In this paper, we make three main contributions.

\textbf{\textit{First,}} to the best of our knowledge, we are the first to formally study the dataset condensation task for diffusion models, a new challenging problem setting that aims at constructing a ``condensed'' sub-dataset with significantly fewer samples than the original dataset for training high-quality diffusion models significantly faster. We address a fundamental academic gap concerning the application of dataset condensation in diffusion models. More specifically, our explorations with the diffusion model provide the first insights into the challenges and potential solutions for applying dataset condensation to vision generation tasks. We note that while conventional dataset condensation has made great progress and sometimes uses diffusion models to construct a subset, this line of research only focused on training discriminative models instead of generative models.

\textit{\textbf{Second}}, we propose \textit{D$^2$C}, a novel two-stage dataset condensation framework tailored for training diffusion models. Our framework addresses the challenges of dataset condensation for diffusion models by decomposing the problem into two key aspects: the \textit{Select} stage identifies an informative, compact, and learnable subset by ranking samples using the diffusion difficulty score derived from a pre-trained diffusion model; the \textit{Attach} stage enriches each selected sample by adding semantic and visual representations, further enhancing the training efficiency while preserving performance.

\textbf{\textit{Third,}} extensive experiments demonstrate great empirical success that the proposed \textbf{\textit{D\textsuperscript{2}C}} can train diffusion models significantly faster with dramatically fewer data while retaining high visual quality, substantiating the effectiveness and scalability. Specifically, \textit{D\textsuperscript{2}C} significantly outperforms random sampling and several popular dataset condensation algorithms across data compression ratios of 0.8\%, 4\%, and 8\%, at resolutions of 256$\times$256 and 512$\times$512, and with both SiT~\citep{SiT} and DiT~\citep{DIT} architectures. In particular, \textit{D\textsuperscript{2}C} achieves a FID of 4.3 in merely 40k training steps using SiT-XL/2~\citep{SiT}, demonstrating a \textbf{100$\times$} acceleration over REPA~\citep{REPA} and a \textbf{233$\times$} speed-up compared to vanilla SiT. Furthermore, it improves to a FID of 2.7 using only 50k condensed images with CFG (refer to Fig.~\ref{fig:first} (c)).

%% file: sec/2_Preliminaries.tex
\section{Preliminaries and Related Work}
\label{diffusion_premi}

\vspace{-4pt}
\noindent{\bf Diffusion Models.} We briefly introduce the standard latent-space noise injection formulation~\citep{DIT}, which defines a forward process that gradually perturbs input data $\mathbf{x}_0 \sim q_0(\mathbf{x})$ with Gaussian noise:
\begin{equation}
q_t(\mathbf{x}_t \mid \mathbf{x}_0) = \mathcal{N}(\mathbf{x}_t ; \alpha_t \mathbf{x}_0, \sigma_t^2 \mathbf{I}),
\end{equation}
where $\alpha_t,\ \sigma_t \in \mathbb{R}^{+}$ are differentiable functions of $t$ with bounded derivatives. The choice for $\alpha_t$ and $\sigma_t$ is referred to as the noise schedule of a diffusion model. After that, we need to train a neural network $\epsilon_\theta(\cdot,\cdot,\cdot)$ to approximate the reverse denoising process (\textit{i.e.}, predict the added noise $\epsilon$) for \revise{sampling} (see Appendix~\ref{AppendixA} for more details). The training objective is to minimize the mean squared error between the predicted and the ground true noise:
\begin{equation}
\label{L_original}
\mathcal{L}_{\text{diff}} = \mathbb{E}_{\mathbf{x}_0\sim q_0(\mathbf{x}), \epsilon\sim\mathcal{N}(0,\mathbf{I}), t\sim \mathcal{U}[0,1]} \left[ \| \epsilon - \epsilon_\theta(\mathbf{x}_t, t, \mathbf{c}) \|^2_2 \right],
\end{equation}
Here, $\mathbf{c}$ is a conditional input, such as class labels or text embeddings. \Rui{In some cases, the prediction target is replaced with the $v$-prediction, which corresponds to flow matching.}

% ~\citep{iclr22_rect,SD35,SDV3}:}
% \begin{equation}
% \mathcal{L}_{\text{velocity}} = \mathbb{E}_{\mathbf{x}_0\sim q_0(\mathbf{x}), \epsilon\sim\mathcal{N}(0,\mathbf{I}), t\sim \mathcal{U}[0,1]} \left[ \| \mathbf{v}_\theta(\mathbf{x}_t, t) -(\epsilon - \mathbf{x}_0 ) \|^2_2 \right].
% \end{equation}
% \Rui{Here, $\mathbf{v}_\theta(\mathbf{x}_t, t)$ denotes the learned time-dependent velocity field.}

\begin{figure*}[t]
  \centering
  \includegraphics[width=0.85\linewidth]{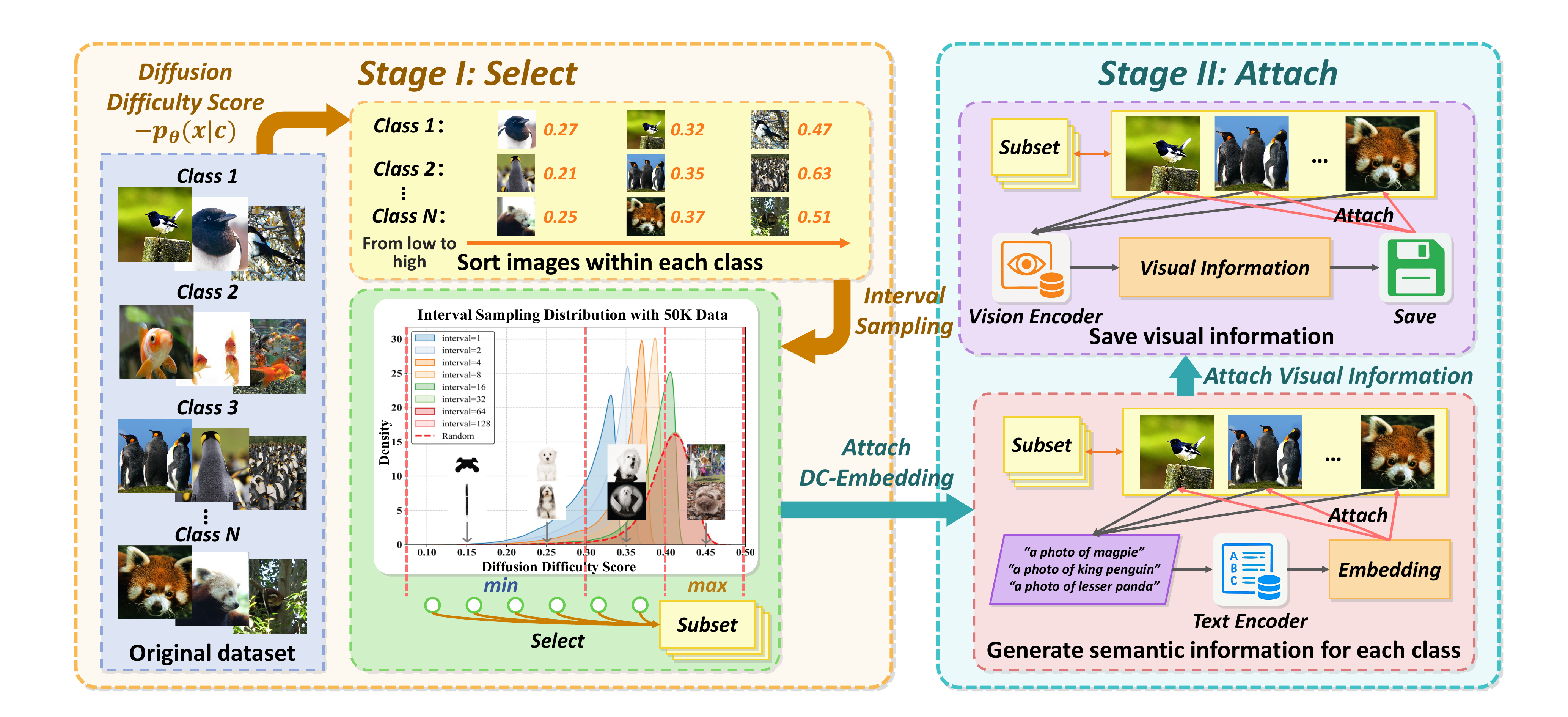}
  \vspace{-6pt}
  \caption{Overview of \textbf{D}iffusion \textbf{D}ataset \textbf{C}ondensation (\textit{\textbf{D$^2$C}}). \textit{D\textsuperscript{2}C} employs a two-stage process: \textit{Select} and \textit{Attach}. The \textit{Select} stage identifies a compact and diverse subset by interval sampling using the diffusion difficulty score derived from a pre-trained diffusion model. The \textit{Attach} stage further enriches each selected sample by adding semantic information and visual information.}
  \label{fig:total_frame work}
  \vspace{-15pt}
\end{figure*}

% The forward process can also be interpreted as a stochastic differential equation (SDE) of the form $\mathrm{d}\mathbf{x}_t = f(t)\mathbf{x}_t\, \mathrm{d}t + g(t)\, \mathrm{d}\mathbf{w}_t$, where $\mathbf{w}_t$ denotes the standard Wiener process, although we focus on the discrete-time formulation in this work.

\noindent{\bf Data-centric Efficient Training.}
\revise{Beyond model-side improvements, a complementary line of work takes a data-centric view on diffusion efficiency.
Patch-based schemes~\citep{ding2023patched,wang2023patch} and Infobatch~\citep{qin2023infobatch} focus on reallocating training effort over existing samples by reweighting or resampling informative regions and instances.
However, comparatively few methods directly tackle diffusion training efficiency by explicitly reducing and restructuring the overall training set.
In this setting, given an original dataset $\mathcal{D} = \{ (\hat{\mathbf{x}}_i, \hat{y}_i) \}_{i=1}^{|\mathcal{D}|}$, where each $\hat{y}_i$ is the label corresponding to sample $\hat{\mathbf{x}}_i$, dataset compression aims to reduce the size of training data while preserving model performance.
Two primary strategies have been extensively studied in this context: dataset pruning and dataset condensation.}

\noindent{\textit{1) Dataset Pruning.}} Dataset pruning selects \revise{an information-enriched} subset from the original dataset, i.e., $\mathcal{D}^\text{core} \subset \mathcal{D}$ with $|\mathcal{D}^\text{core}| \ll |\mathcal{D}|$, and directly minimizes the training loss over the subset:
\begin{equation}
\min_\theta \mathbb{E}_{(\mathbf{x}, y) \sim \mathcal{D}^\text{core}} \left[ \ell(\phi_{\theta_{\mathcal{D}^\text{core}}}(\mathbf{x}), y) \right],
\end{equation} 
where $\ell(\cdot,\cdot)$ denotes the empirical training loss, and $\phi_{\theta_{\mathcal{D}^\text{core}}}$ is the model parameterized by $\theta_{\mathcal{D}^\text{core}}$. Classical data pruning methods like random sampling, K-Center~\citep{kcenter}, and Herding~\citep{herding} can be used with diffusion models, but they offer minimal performance improvements.
\revise{Very recently, Li et al.~\citep{li2025pruning} investigate data-efficient diffusion training from the perspective of dataset pruning by selecting a coreset with surrogate features and then performing class-wise reweighting. While this approach substantially reduces training cost and improves over naive pruning, it does not attach any additional information to the selected samples and is mainly validated on relatively small-scale or latent diffusion settings, which limits its ability to fully exploit the potential of condensed training data for large-scale, high-resolution diffusion models.}

\noindent{\textit{2) Dataset Condensation.}} 
\RuiR{Following recent work~\citep{wu2025dc3}, dataset condensation aims to synthesize a small, compact, and diverse synthetic dataset $\mathcal{D}^\mathcal{S} = (\mathbf{X}, \mathbf{Y}) = \{ (\mathbf{x}_j, y_j) \}_{j=1}^{|\mathcal{D}^\mathcal{S}|}$ to replace the original dataset $\mathcal{D}$. The synthetic dataset $\mathcal{D}^\mathcal{S}$ is generated by a condensation algorithm $\mathcal{C}$ such that $\mathcal{D}^\mathcal{S} \in \mathcal{C}(\mathcal{D})$, with $|\mathcal{D}^\mathcal{S}| \ll |\mathcal{D}|$. Each $y_j$ corresponds to the synthetic label for the sample $\mathbf{x}_j$.

The key motivation for dataset condensation is to create $\mathcal{D}^\mathcal{S}$ such that models trained on it can achieve performance within an acceptable deviation $\eta$ compared to models trained on $\mathcal{D}$. This can be formally expressed as:
\begin{equation}
\label{equ:dc}
\sup \left\{ \left| \ell(\phi_{\theta_\mathcal{D}}(\hat{\mathbf{x}}), \hat{y}) - \ell(\phi_{\theta_\mathcal{D}^\mathcal{S}}(\hat{\mathbf{x}}), \hat{y}) \right| \right\}_{(\hat{\mathbf{x}}, \hat{y}) \sim \mathcal{D}} \leq \eta,
\end{equation}
where $\theta_\mathcal{D}$ is the parameter set of the neural network $\phi$ optimized on $\mathcal{D}$:
$\theta_\mathcal{D} = \arg\min_\theta \mathbb{E}_{(\hat{\mathbf{x}}, \hat{y}) \sim \mathcal{D}} \left[ \ell(\phi_\theta(\hat{\mathbf{x}}), \hat{y}) \right].$
A similar definition applies to $\theta_\mathcal{D}^\mathcal{S}$, which is optimized on the synthetic dataset $\mathcal{D}^\mathcal{S}$.} Existing DC methods can be broadly divided into two families. \revise{Pixel-level} approaches perform dataset distillation by directly optimizing synthetic training images in pixel space (e.g., using gradient- or matching-based objectives)~\citep{dd_sre2l,dd_mtt,EDC,shao2023generalized}. In contrast, \revise{image-level} condensation operates on real images via selection and transformation, as in patch-based or quantization-style schemes~\citep{RDED,wu2025dc3,OD3}. These methods have been developed mainly for discriminative models; when naively applied to diffusion training, they tend to produce images that deviate from the target data distribution, which harms generative quality (see Appendix~\ref{appdix:concendation} for visualizations). \revise{Our \textit{D\textsuperscript{2}C} framework follows the image-level condensation route, but goes beyond passive pruning by not only selecting informative real samples, but also attaching rich semantic and visual representations tailored to diffusion training.}

%% file: sec/3_method.tex
\section{Diffusion Dataset Condensation} %(\textbf{D\textsuperscript{2}C})

% To enable data-centric efficient training of diffusion models under limited resources, we propose \textbf{D}iffusion \textbf{D}ataset \textbf{C}ondensation (\textbf{\textit{D\textsuperscript{2}C}}), the first unified framework that systematically condenses training data for diffusion models.  this process produces a condensed dataset suitable for efficient diffusion model training.

As illustrated in Fig.~\ref{fig:total_frame work}, {\textit{D\textsuperscript{2}C}} consists of two stages: \textit{Select} (Sec.~\ref{section_select}), which identifies a compact set of diverse and learnable real images using \textit{diffusion difficulty score} and \textit{interval sampling} techniques; and \textit{Attach} (Sec.~\ref{section_attach}), which augments each selected image with semantic and visual information to improve generation performance. Finally, we describe how to train diffusion models on the condensed dataset produced by \textit{D\textsuperscript{2}C} in Sec.~\ref{section_training}. (Sec.~\ref{section_training}).

% \xie{I prefer to present Figure 2 here, where you first refer it.}

\subsection{\textit{Select}: Difficulty-Aware Selection}
\label{section_select}

In this work, we focus on class-to-image (C2I) synthesis, aligned with the setting in \citet{REPA}, and show that our framework also applies to the text-to-image (T2I) setting with only minor changes; see Appendix~\ref{sec:select_t2i} for details. Given a class-conditioned dataset $\mathcal{D} = \bigcup_{y=1}^{C} \mathcal{D}_y$, where $C$ denotes the class number and $\mathcal{D}_y= \{x_i\}_{i=1}^{|\mathcal{D}_y|}$ denotes all samples of class $y$, our aim is to select a compact subset for efficient diffusion training. To achieve this, we propose the \textit{ diffusion difficulty score} to quantify the denoising difficulty of each sample, followed by our designed \textit{interval sampling} to ensure diversity within the selected subset.

\noindent{\bf Diffusion Difficulty Score.} The arrangement of samples from easy to hard is crucial for revealing underlying data patterns and facilitating difficulty-aware selection.
Recent work~\citep{li2023your, zipengsimple} demonstrates that diffusion models inherently encode semantic-related class-conditional probability $p_\theta(\mathbf{c}\mid\mathbf{x})$ through the variational lower bound (i.e., diffusion loss Eq.~\ref{L_original}) of $\log p_\theta(\mathbf{x}\mid\mathbf{c})$~\citep{ddpm_begin,sde}. This conditional probability admits the standard Bayesian form
\begin{equation}
\label{eq:bayes_posterior}
p_\theta(\mathbf{c}\mid\mathbf{x})
= \frac{p_\theta(\mathbf{x}\mid\mathbf{c})\,p(\mathbf{c})}
       {\sum_{\hat{c}} p_\theta(\mathbf{x}\mid\hat{c})\,p(\hat{c})}.
\end{equation}
Intuitively, a larger $p_\theta(\mathbf{c}\mid\mathbf{x})$ indicates that sample $\mathbf{x}$ can be more confidently identified as belonging to class $\mathbf{c}$, thus suggesting lower learning difficulty. Computing the full denominator in Eq.~\eqref{eq:bayes_posterior} for every sample is expensive, while we only need a score that orders samples by difficulty. Since the class label $y \sim U\{1,\dots,C\}$ is obtained by uniform sampling and the average likelihood over classes does not vary too much across samples, i.e., we assume
$\sup_{\mathbf{x}_1,\mathbf{x}_2\sim\mathcal{D}} \bigl|\mathbb{E}_{\hat{c}}[p_\theta(\mathbf{x}_1\mid\hat{c})] - \mathbb{E}_{\hat{c}}[p_\theta(\mathbf{x}_2\mid\hat{c})]\bigr| \le \eta$,where $\eta>0$ is a small tolerance,
the denominator in Eq.~\eqref{eq:bayes_posterior} can be treated as approximately constant with respect to $\mathbf{x}$. Consequently, the posterior is proportional to the class-conditional likelihood,
\begin{equation}
\label{eq:posterior_propto}
p_\theta(\mathbf{c}\mid\mathbf{x}) \propto p_\theta(\mathbf{x}\mid\mathbf{c}).
\end{equation}

We define the diffusion difficulty score based on this posterior:
\begin{equation}
\label{eq-sdiff}
\begin{aligned}
s_\text{diff}(\mathbf{x})
&= -p_\theta(\mathbf{c}\mid\mathbf{x})
   \propto -p_\theta(\mathbf{x}\mid\mathbf{c}) \\
&= 
-\mathbb{E}_{\epsilon\sim\mathcal{N}(0,\mathbf{I}),\,t\sim\mathcal{U}[0,1]}
 \left[\left\| \epsilon - \epsilon_\theta\left( \mathbf{x_t} ,  t, \mathbf{c} \right) \right\|^2_2\right]
\end{aligned}
\end{equation}

The higher the score $s_\text{diff}(\mathbf{x})$, the more difficult it is, and the lower the score $s_\text{diff}(\mathbf{x})$, the easier it is. To simplify our presentation, we define the diffusion loss $ - p_\theta(\mathbf{x}|\mathbf{c})$ as the diffusion difficulty score.

\begin{figure*}[t]
  \centering
  \includegraphics[width=0.95\linewidth]{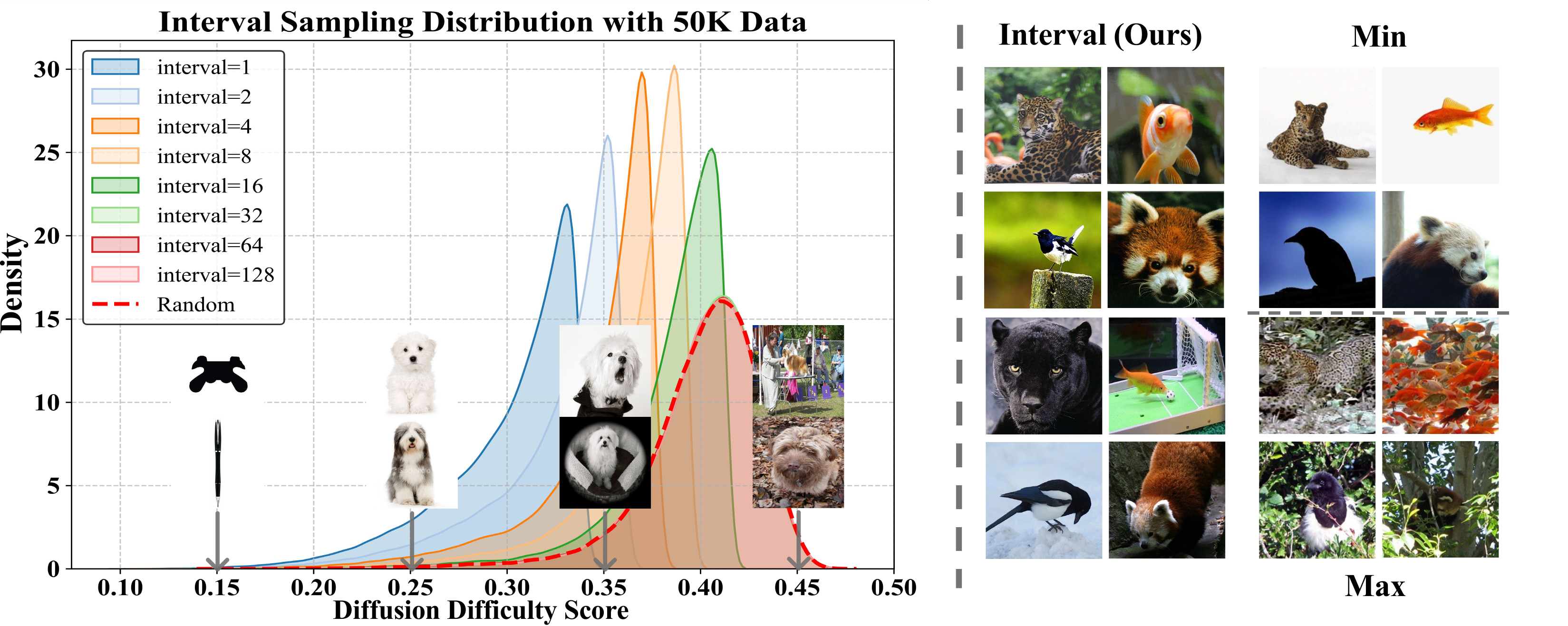}
  \vspace{-10pt}
  \caption{\textbf{Left:} Distribution of diffusion difficulty scores under different interval values $k$. Smaller intervals (e.g., 1, 2) favor low-loss samples, while larger intervals (e.g., 64, 128) result in a distribution closer to random sampling, thus approximating the original data distribution. Moderate intervals (e.g., 16) provide balanced coverage across difficulty levels. \textbf{Right:} Representative samples selected by three strategies: \textit{Min} (lowest score), \textit{Max} (highest score), and \textit{Interval} (our proposed strategy). Interval sampling achieves a balance between structural clarity and contextual richness.}
    \vspace{-10pt}
  \label{fig:loss_vis}
\end{figure*}

% \RuiR{The min and max need to be unified in both the figure and the presentation to prevent confusion. We can discuss this point}
By computing $s_\text{diff}(x)$ for all training samples, we construct a ranked dataset. As shown in Fig.~\ref{fig:loss_vis}, these scores exhibit a skewed unimodal distribution. Selecting the easiest samples (\textit{Min}) yields a subset dominated by clean, background-simple images with high learnability but limited diversity. In contrast, selecting only the highest-score samples (\textit{Max}) results in cluttered, noisy, and ambiguous images that are difficult to optimize. Meanwhile, many samples lie in the middle range, offering moderate learnability but richer contextual information. Selecting an appropriate value within this range is therefore critical; we provide a more detailed discussion in Appendix~\ref{app:dds_algo}.

\noindent{\bf Interval Sampling.} To balance diversity and learnability, we propose an \textit{interval sampling} strategy. Specifically, we sort its images $\mathcal{D}_y$ within each class $y$ in ascending order of $s_\text{diff}(x)$ and select samples at a fixed interval $k$: $\mathcal{D}_{\text{IS}} = \bigcup_{y=1}^{C} \left\{ x^{(i)} \in \mathcal{D}_y \;\middle|\; i \in \{0, k, 2k, \dots \} \right\}$, where $\mathcal{D}_{\text{IS}}$ denotes the selected subset constructed by interval sampling, $k$ is the fixed sampling interval, and $x^{(i)}$ is the $i$-th sample in the sorted list (e.g., $x^{(0)}$ corresponds to the sample with the lowest diffusion difficulty score). Interval sampling with a larger interval $k$ promotes diversity in the sampled data while potentially hindering learnability. As shown in Fig.~\ref{fig:loss_vis} (Left), this trade-off arises from a shift in the sample distribution: a larger $k$ leads to a reduction in the number of easy samples and a corresponding increase in the representation of standard and difficult samples.

% When the selected subset does not reach the desired quota due to large $k$ or insufficient samples, we supplement the remaining selection by randomly sampling from $\mathcal{D}_y$, resulting in a subset that is primarily composed of randomly selected samples and thus approximates the original data distribution.

\begin{figure}[t]
    \vspace{-5pt}
    \centering
    \includegraphics[width=0.8\linewidth]{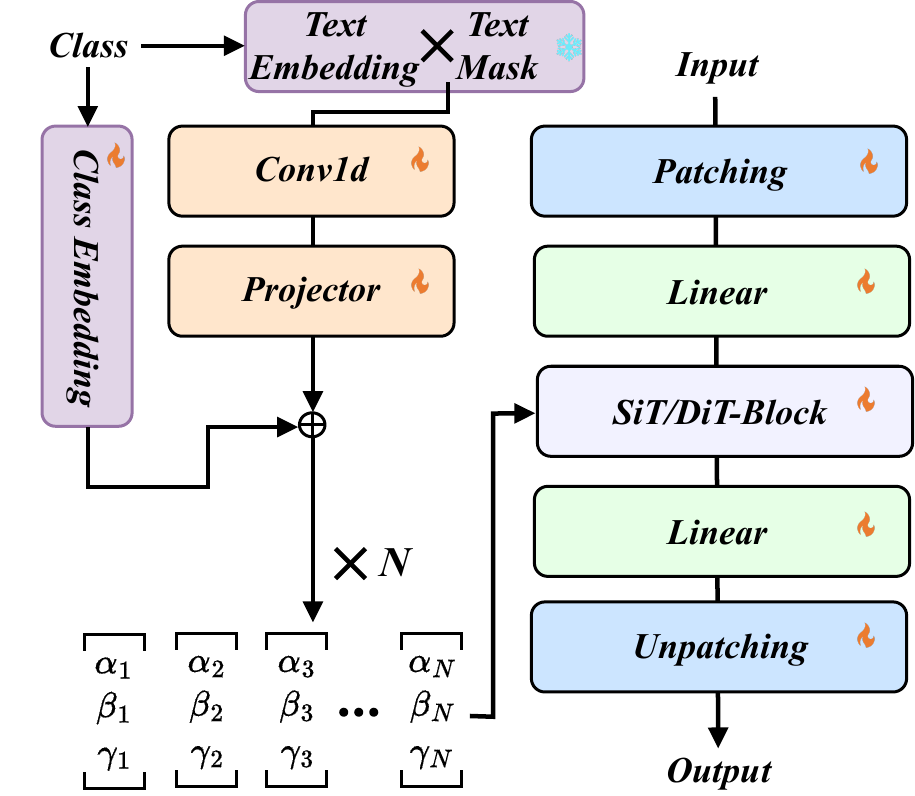}
    \caption{Overview of DC-Embedding.}
    \label{fig:text-embedding}
    \vspace{-13pt}
\end{figure}

\noindent{\bf Extended Discussion.} Training exclusively on the easiest (Min) or the hardest (Max) samples is suboptimal. Instead, a balanced curriculum comprising easy, medium, and difficult examples yields a training subset that is both learnable and diverse, ultimately leading to stronger generative performance. We further offer more discussions and insights on interval sampling in Appendix~\ref{app:select_insights}.

\subsection{\textit{Attach}: Semantic and Visual Information Enhancement}
\label{section_attach}

To complement the \textit{Select} phase, which yields a compact subset of informative real images, the \textit{Attach} phase enriches each selected instance with additional semantic and visual information. In particular, we attach semantic information via a Dual Conditional Embedding (DC-Embedding) module and inject visual information through  visual representation, resulting in a more expressive condensed dataset and improved generalization of the trained diffusion models.

\noindent{\bf Dual Conditional Embedding (DC-Embedding).} Existing C2I synthesis methods~\citep{DIT,SiT} commonly rely on class embeddings trained from scratch, which often fail to effectively capture inherent semantic information (see Appendix~\ref{appendix:DC embedding}). We enrich the class embedding by incorporating text representations derived from a pre-trained text encoder (e.g., T5-encoder~\citep{ni2021sentence}). For each class $c \in \{1, \dots, C\}$, a descriptive prompt $P(c)$ (e.g., \textit{``a photo of a cat''}) is encoded by a pre-trained text encoder $f_\text{text}$, yielding its corresponding text embedding $t_c$ and text mask $t_\textrm{mask}$:
\begin{equation}
    \label{eq:txet1}
    \begin{split}
        t_c,t_\textrm{mask} &= f_\text{text}(P(c)), \\
    \end{split}
\end{equation}
The resulting text embedding and text mask are stored on disk as attached text information alongside the subset $\mathcal{D}_\textrm{IS}$ generated in the preceding phase, ready for import during formal training. During the formal training, as illustrated in Fig.~\ref{fig:text-embedding}, the text embedding $t_c$ and the text mask $t_\textrm{mask}$ undergo a 1D convolution and are fused with a learnable class embedding $e_c$ using a residual MLP:\begin{equation}
\begin{aligned}
\label{eq:txet2}
\tilde{t}_c &= \text{Conv1d}(t_c\times t_\textrm{mask}),\quad y_\text{text} = \text{MLP}(\tilde{t}_c) + \tilde{t}_c + e_c.
\end{aligned}
\end{equation}

\begin{figure*}[t]
  \centering
  \includegraphics[width=0.99\linewidth]{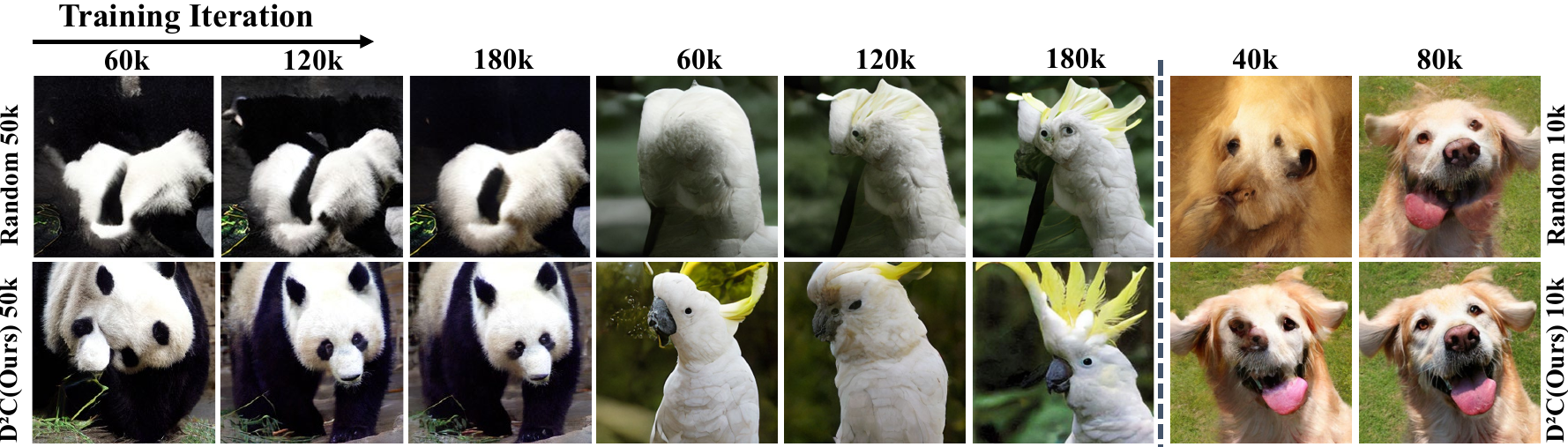}
    \vspace{-6pt}
    \caption{\textbf{\textit{D$^2$C} improves visual quality under tight data budgets.} We compare Random sampling and \textit{D$^2$C} on DiT-L/2 at 10k and 50k data budgets, and neither setting uses classifier-free guidance.}
  \label{fig:visualization}
    \vspace{-14pt}
\end{figure*}

This resulting vector $y_\text{text}$ then serves as a semantic conditioning token for the conditional diffusion model. Compared to using simple class embeddings alone, this formulation offers richer semantic information while retaining the learnability of class embeddings.

\noindent{\bf Visual Information Injection.} While semantic information aids in distinguishing inter-class structure, it often fails to capture the intra-class variability essential for high-fidelity generation. To address this, we integrate instance-specific visual representations into the attached information. For each image $x \in \mathbb{R}^{3 \times H \times W}$, a pre-trained vision encoder $f_\text{vis}$ (e.g., DINOv2~\citep{dinov2}) extracts patch-level semantic representations:
\begin{equation}
\label{equ:vis}
y_\text{vis} = f_\text{vis}(x) \in \mathbb{R}^{N \times d_\text{text}}
\end{equation}
where $N$ is the number of image patches and $d_\text{text}$ is the feature dimension. We retain the first $h$ (i.e., number of tokens in the diffusion transformer) tokens of $y_\text{vis}$ to form a compact representation of the dominant structure: $y_\text{vis} = y_\text{vis}[:\!h,:] \in \mathbb{R}^{h \times d_\text{text}}$. As outlined in REPA~\citep{REPA}, this visual information provides a semantic prior for the diffusion model and thus significantly benefits data-centric efficient training. Similar to the text information $y_\text{text}$, the visual information $y_\text{vis}$ is also stored on disk as attached metadata alongside the selected subset $\mathcal{D}_\text{IS}$.

% . In the subsequent training stage, we align the model’s intermediate representations with these extracted features (see Sec.~\ref{sec:training}), effectively guiding the network to retain class-consistent structure while capturing intra-class diversity.

\subsection{\textit{D\textsuperscript{2}C} Training Process}
\label{section_training}
Here, we detail the training process of the diffusion model using our condensed dataset, which comprises a compact subset selected during the \textit{Select} phase and subsequently enriched with semantic and visual information during the \textit{Attach} phase. Our goal is to fully leverage the information contained in our condensed dataset to accelerate training without compromising performance.

We employ a conditional diffusion model $\mathcal{D}_\theta$ and, as an example, utilize the optimization objective of score-based diffusion models: predicting the added noise $\epsilon$ from the perturbed latent input $\mathbf{x}_t$ at time step $t$, conditioned on the text information $y_\textrm{text}$ and the class label $y$. The new denoising loss is defined as $\mathcal{L}_{\text{diff}} = \mathbb{E}_{\mathbf{x}_0\sim q_0(\mathbf{x}), \epsilon\sim\mathcal{N}(0,\mathbf{I}), t\sim \mathcal{U}[0,1]} \left[ \| \epsilon - \epsilon_\theta(\mathbf{x}_t, t, y, y_\textrm{text}) \|^2_2 \right]$, where the specific injected forms of $y$ and $y_\text{text}$ can be found in Sec.~\ref{section_attach}. Then, to maximize the utilization of visual information, we adopt the same formulation as REPA~\citep{REPA}, which involves aligning the encoder's output (i.e., the decoder's input) within the diffusion model with the visual representation $y_\text{vis} = \{v_i\}_{i=1}^h$. Concretely, from a designated intermediate layer of the diffusion backbone, we obtain token features $\{ h_i \in \mathbb{R}^d \}_{i=1}^h$. A projection head $\phi$ maps these tokens from $\mathbb{R}^d$ to $\mathbb{R}^{d_\text{text}}$, and we compute a semantic alignment loss:\begin{equation}
\mathcal{L}_{\text{proj}} = - \frac{1}{h} \sum_{i=1}^h \left\langle \frac{\phi(h_i)}{\| \phi(h_i) \|}, \frac{v_i}{\| v_i \|} \right\rangle.
\end{equation}This loss encourages the model to align its encoder's output with visual representations, promoting localized realism and spatial consistency~\citep{dinov2} in generation.

\noindent{\bf Overall Training Objective.}
The final training loss combines the denoising objective and the semantic alignment term (with the balance weight $\lambda$ is set to 0.5 by default):
\begin{equation}
\mathcal{L}_{\text{total}} = \mathcal{L}_{\text{diff}} + \lambda \mathbb{E}_{\mathbf{x},\epsilon\sim\mathcal{N}(0,\mathbf{I}),t\sim\mathcal{U}[0,1],y,y_\text{text},y_\text{vis}}\left[\mathcal{L}_{\text{proj}}\right].
\end{equation}
This training strategy enables \textit{D\textsuperscript{2}C} to effectively learn from limited yet enhanced data, offering a practical solution for efficient diffusion training under minimal budgets.
\definecolor{lightpurple}{RGB}{230,210,255}
\definecolor{lightblue}{RGB}{204,229,255}
\definecolor{lightgreen}{RGB}{220,255,220}
\definecolor{lightyellow}{RGB}{255,245,200}

%% file: sec/4_experiment.tex
\section{Experiments}
\label{sec:experiment}

In this section, we validate the performance of \textit{D\textsuperscript{2}C} and analyze the contributions of its components through extensive experiments. In particular, we aim to answer the following questions: \textcolor{C3}{\textit{\textbf{1)}}}\ Can \textit{D\textsuperscript{2}C} improve training speed and reduce data usage of diffusion models? \noindent\textcolor{C3}{\textit{\textbf{2)}}}\ Does \textit{D\textsuperscript{2}C} generalize well across backbones, data scales, and resolutions? \noindent\textcolor{C3}{\textit{\textbf{3)}}}\ How do \textit{D\textsuperscript{2}C}'s components and hyperparameter choices affect its overall effectiveness? 

% (Fig.~\ref{fig:loss_vis},~\ref{fig:interval&componets_ablation})

% (Fig.~\ref{fig:loss_vis},~\ref{fig:interval&componets_ablation},~\ref{fig:interval_ablation})

\begin{table*}[t]
\caption{Comparison of gFID-50K across various dataset condensation methods and data budgets using DiT-L/2 and SiT-L/2 on ImageNet 256$\times$256. We use CFG=1.5 for evaluation. \textit{D$^2$C} surpasses other methods at all settings.}
\label{tab:main_comparison}
\vspace{-5pt}
\centering
\footnotesize
\setlength{\tabcolsep}{5pt} % 调整列间距
\renewcommand{\arraystretch}{0.9} % Adjust this value to 
\begin{tabular}{cc|ccc|a|ccc|a}
\toprule
\multirow{2}{*}{Data Budget} & \multirow{2}{*}{Iter.} & \multicolumn{4}{c|}{DiT-L/2} & \multicolumn{4}{c}{SiT-L/2} \\
 & & Random & K-Center & Herding & \textit{D$^2$C} & Random & K-Center & Herding & \textit{D$^2$C} \\
\midrule
0.8\% (10K) & 100k & 35.86 & 50.77 & 40.75 & \textbf{4.20}  & 4.35  & 14.77 & 22.96 &  \textbf{3.98} \\
0.8\% (10K) & 300k & 4.19  & 13.5  & 22.35 & \textbf{4.13} & 4.33  & 13.58 & 22.55 &  \textbf{3.98} \\
\midrule
4.0\% (50K) & 100k & 36.78 & 69.86 & 32.38 & \textbf{14.81}& 31.13 & 61.66 & 29.11 &  \textbf{11.21} \\
4.0\% (50K) & 300k & 11.55 & 38.54 & 22.44 & \textbf{5.99}& 14.18 & 39.69 & 22.44 &  \textbf{5.66} \\
\midrule
8.0\% (100K) & 100k & 41.02 & 71.31 & 36.37 & \textbf{   22.55   }& 36.64 & 66.96 & 32.3  &  \textbf{15.01} \\
8.0\% (100K) & 300k & 11.49 & 37.35 & 15.23 & \textbf{  6.49    }& 12.56 & 39.08 & 16.17 &  \textbf{5.65} \\
\bottomrule
\end{tabular}
% }
\label{tab:ipc_comparison}
\vspace{-8pt}
\end{table*}

\subsection{Setup}

\noindent\textbf{Experiment settings.} We conduct experiments on the ImageNet-1K dataset~\citep{ILSVRC15}, using subsets of 10K, 50K, and 100K images, corresponding to 0.8\%, 4\%, and 8\% of the full dataset, respectively. To further demonstrate the generalization and effectiveness of our method, Appendix~\ref{appdix:cifar} reports additional results of \textit{D\textsuperscript{2}C} on CIFAR datasets. All images are center-cropped and resized to 256$\times$256 and 512$\times$512 resolutions using the ADM~\cite{adm2021} preprocessing pipeline. Furthermore, we use [$\cdot$]-L/2 and [$\cdot$]-XL/2 architectures in both DiT~\cite{DIT} and SiT~\cite{SiT} backbones, following the standard settings outlined in~\citet{SiT}.

\begin{table}[t]
\caption{Comparison with a strict data budget 0.8\% (10K) on ImageNet 512$\times$512. We use CFG=1.5 for evaluation. \textit{D$^2$C} surpasses random sampling at all settings.}
\label{tab:512Xcomparision}
\centering
\vspace{-5pt}
\footnotesize
\setlength{\tabcolsep}{4.4pt} % 调整列间距
\renewcommand{\arraystretch}{0.9} % Adjust this value to 

\resizebox{\linewidth}{!}{%  ← 新增这一行
\begin{tabular}{lllcccc}
\toprule
Model & Method & Iter. & gFID$\downarrow$ & sFID$\downarrow$ & {Inception Score}$\uparrow$ & {Precision}$\uparrow$ \\
\midrule
DiT-L/2        & Random & 100k    & 24.8 & 11.9 & 74.3  & 0.65 \\
\CC {DiT-L/2}        & \CC \textit{D\textsuperscript{2}C} (Ours) & \CC 100k    & \CC \textbf{14.8} & \CC \textbf{6.9} & \CC \textbf{109.2}  & \CC \textbf{0.63} \\
\cmidrule(lr){1-7}
DiT-L/2        & Random & 300k    & 17.1 & \textbf{12.8} & 130.6 & 0.64 \\
\CC {DiT-L/2}        & \CC {\textit{D\textsuperscript{2}C} (Ours)} & \CC {300k}    & \CC \textbf{5.8} & \CC {15.1} & \CC \textbf{318.9} & \CC \textbf{0.77} \\
\midrule
SiT-L/2        & Random & 100k    & 13.3 & 22.8 & 197.1 & 0.69 \\
\CC {SiT-L/2}        & \CC {\textit{D\textsuperscript{2}C} (Ours)} & \CC{100k}    & \CC \textbf{9.1} & \CC \textbf{14.3} & \CC \textbf{261.7} & \CC \textbf{0.72} \\
\cmidrule(lr){1-7}
SiT-L/2        & Random & 300k    & 5.0  & 13.6 & \textbf{316.9} & 0.76 \\
\CC {SiT-L/2}        & \CC {\textit{D\textsuperscript{2}C} (Ours)} & \CC {300k}    & \CC \textbf{4.22}  & \CC \textbf{11.6} & \CC 289.7 & \CC \textbf{0.79} \\
\bottomrule
\end{tabular}
}% ← 这里补上对应的 }
\vspace{-20pt}
\end{table}

\noindent\textbf{Evaluation and baselines.} We train models from scratch on the collected subset and evaluate them using gFID~\citep{fid}, sFID, Inception Score~\citep{is} and Precision, adhering to standard evaluation protocols~\citep{adm2021,DIT,SiT}. We compare our method against REPA~\citep{REPA}, REPA-E~\citep{REPA-E}, REG~\citep{REG} and various data condensation and selection baselines, including SRe$^2$L~\cite{dd_sre2l}, RDED~\citep{RDED}, Herding, K-Center, and random sampling, using SiT and DiT architectures~\citep{SiT,DIT}. Further details regarding evaluation metrics and baseline methods can be found in Appendix~\ref{evaluation_details} and~\ref{Baeline_details}.

\begin{figure*}[t]
  \centering
  \includegraphics[width=1.0\linewidth]{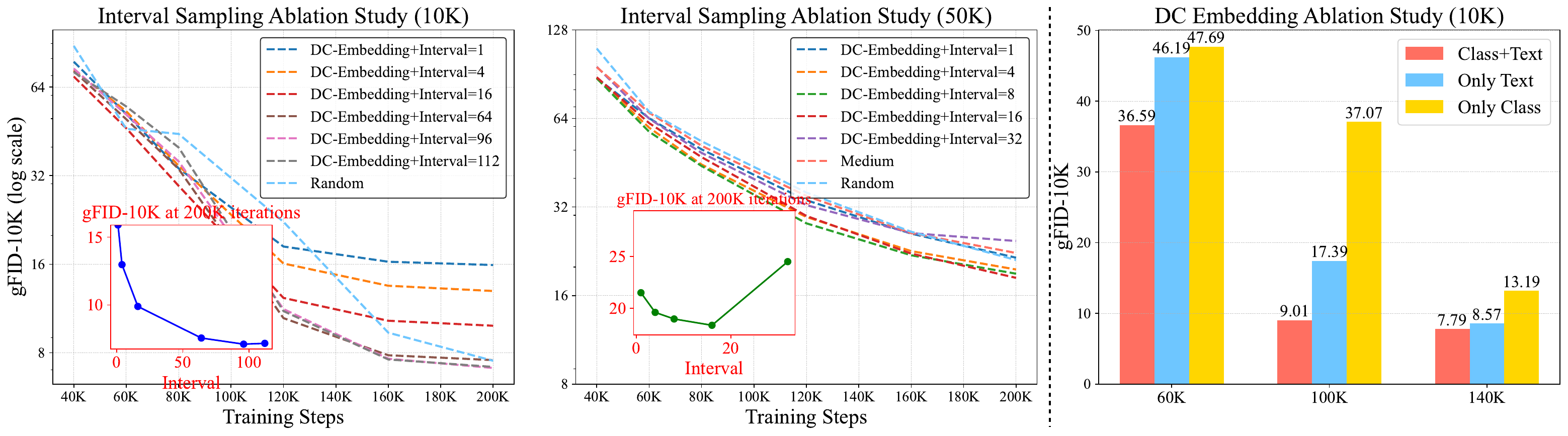}
  \vspace{-14pt}
    \caption{\textbf{Left: Interval-sampling ablation.}  Small $k$ speeds early training. The best final gFID-10K appears at $k{=}96$ for the 10K budget and $k{=}16$ for the 50K budget, roughly scaling with data size. \textbf{Right: DC-Embedding ablation at 10K.} Combining text and class embeddings outperforms either alone; ``Only Class'' denotes the baseline that injects class embeddings only.}
  \label{fig:interval&componets_ablation}
  \vspace{-5pt}
\end{figure*}

\vspace{-5pt}
\subsection{Main Result}

\noindent{\bf Training Performance and Speed.} We evaluate \textit{D$^2$C} using 10K and 50K data budgets, comparing its performance against REPA and a vanilla SiT model trained on the full ImageNet dataset (a 1.28M data budget), as well as random selection with 10K and 50K data budgets. As shown in Table~\ref{tab:main} and Fig.~\ref{fig:first} (b), our method achieves a gFID-50K of 4.23 at only 40K iterations with 10K training data. In contrast, REPA requires 4 million steps and the vanilla SiT model needs 7 million steps to reach comparable performance, representing an acceleration of over \textbf{100$\times$} and \textbf{233$\times$}, respectively. Under a 4\% data budget (50K) with CFG set to 1.5, our method achieves an FID of 2.78 at 180K steps, further demonstrating significant data and compute efficiency (Fig.~\ref{fig:first} (c)). Moreover, Fig.~\ref{fig:visualization} presents a visual comparison between random selection and our \textit{D$^2$C} at 10K and 50K data sizes. Our method demonstrates superior visual quality compared to the baseline and generates higher-quality images, even during the early iterations of training.

\begin{table}[t]
  \centering
  \small
  \captionsetup{justification=centering,singlelinecheck=false,skip=4pt} 
  \caption{Comparison of acceleration algorithms on ImageNet-1K.}
  \label{tab:main}
  \vspace{-5pt}
  \resizebox{0.85\linewidth}{!}{% 可以改回 \linewidth，看你需要的大小
    \begin{tabular}{lrrr}
      \toprule
      Model & Training Set & Iter. & gFID$\downarrow$ \\
      \midrule
      DiT L-2 & 1.28M & 400k & 23.3 \\
      $+$ REPA & 1.28M & 400k & 15.6 \\
      \CC \bfseries $+$ \textit{D\textsuperscript{2}C} & \CC \textbf{0.05M} & \CC \textbf{10k} & \CC \textbf{14.81} \\
      \CC \bfseries $+$ \textit{D\textsuperscript{2}C} & \CC \textbf{0.01M} & \CC \textbf{10k} & \CC \textbf{4.2} \\
      \midrule[\heavyrulewidth]
      SiT L-2 & 1.28M & 400k & 18.8 \\
      $+$ REPA & 1.28M & 700k & 8.4 \\
      \CC \bfseries $+$ \textit{D\textsuperscript{2}C} & \CC \textbf{0.01M} & \CC \textbf{80k} & \CC \textbf{7.07} \\
      \midrule
      SiT XL-2 & 1.28M & 7M & 8.3 \\
      $+$ REPA & 1.28M & 4M & 5.9 \\
      $+$ REPA-E & 1.28M & 235k & 5.9 \\
      $+$ REG & 1.28M & 200k & 5.0 \\
      \CC \bfseries $+$ \textit{D\textsuperscript{2}C} & \CC \textbf{0.01M} & \CC \textbf{40k} & \CC \textbf{4.3} \\
      \CC \bfseries $+$ \textit{D\textsuperscript{2}C} & \CC \textbf{0.05M} & \CC \textbf{180k} & \CC \textbf{2.78} \\
      \bottomrule
    \end{tabular}
  }
  \vspace{-20pt}
\end{table}

\noindent{\bf Comparison on ImageNet 256$\times$256.} We compare \textit{D\textsuperscript{2}C} with random sampling, Herding~\citep{herding}, K-Center~\citep{kcenter}, SRe$^2$L~\citep{dd_sre2l}, RDED~\citep{RDED} under various data budgets and backbones. As shown in Table~\ref{tab:main_comparison}, \textit{D\textsuperscript{2}C} consistently achieves the lowest FID across all settings. For instance, using only 0.8\% of the data and 100K iterations with early stopping, our method achieves a gFID-50K of 4.20 on DiT-L/2 and 3.98 on SiT. These results demonstrate the superiority of our approach over existing methods. Notably, SRe$^2$L and RDED, which perform well in classification task, fail on this generative task (see Table~\ref{tab:sre2l_comparison}) due to their focus on category-discriminative features. Similarly, geometry-based methods like Herding and K-Center, along with random sampling, prove inadequate for achieving efficient and high-performing training.

\noindent{\bf Comparison on ImageNet 512$\times$512.} As shown in Table~\ref{tab:512Xcomparision}, \textit{D\textsuperscript{2}C} achieves a gFID of 5.8 on DiT-L/2, a significant improvement over the 17.1 achieved by random sampling at 300k iterations under the ImageNet 512$\times$512 settings. On SiT-L/2, similar improvements are observed. These demonstrate that \textit{D\textsuperscript{2}C} generalizes well to higher resolutions.

\vspace{-5pt}
\subsection{Ablation Study}
\label{sec:ablation}

\begin{table}[t]
\vspace{-3pt}
\caption{\textit{D$^2$C} vs. SRe$^2$L~\citep{dd_sre2l} and RDED~\citep{RDED} on ImageNet 256$\times$256 with a data budget 0.8\% (10K). }
\label{tab:sre2l_comparison}
\centering
\vspace{-5pt}
\footnotesize
\setlength{\tabcolsep}{4.4pt} % 调整列间距
\renewcommand{\arraystretch}{0.9} % Adjust this value to 
\begin{tabular}{llcccc}
\toprule
Model & Method & gFID$\downarrow$ & sFID$\downarrow$ & {Inception Score}$\uparrow$ & {Precision}$\uparrow$ \\
\midrule
DiT-L/2 & RDED        & 166.2 & 60.1 & 10.8  & 0.09 \\
DiT-L/2 & SRe$^2$L        & 104.2 & 20.2 & 14.1  & 0.20 \\
\CC {DiT-L/2} & \CC {\textit{D\textsuperscript{2}C} (Ours)} & \CC \textbf{4.2}   & \CC \textbf{11.0} & \CC \textbf{283.6} & \CC \textbf{0.72} \\
\midrule
SiT-L/2 & RDED        & 97.5  & 66.8 & 65.63  & 0.22 \\
SiT-L/2 & SRe$^2$L        & 82.3  & 19.8 & 18.1  & 0.27 \\
\CC {SiT-L/2} & \CC {\textit{D\textsuperscript{2}C} (Ours)} & \CC \textbf{3.9}   & \CC \textbf{10.7} & \CC \textbf{289.7} & \CC \textbf{0.73} \\
\bottomrule
\end{tabular}
\vspace{-16pt}
\end{table}

\textbf{Ablation on \textit{Select} Phase.} We investigate the impact of the interval value $k$ in the \textit{Select} phase, as shown in Fig.~\ref{fig:interval&componets_ablation} (Left). Using a small value accelerates early training by prioritizing min-loss samples, which are simpler and easier to learn. However, the limited diversity of such samples leads to degraded performance in later stages, eventually being overtaken by settings with moderate interval values. In contrast, large intervals or random selection introduce excessive max-loss or uncurated samples, destabilizing training (Fig.~\ref{fig:loss_vis}). As $k$ increases, we observe that gFID-10K first decreases and then worsens, revealing an optimal trade-off between diversity and learnability. Empirically, the best results are achieved with an interval of 96 for the 10K budget and 16 for 50K, approximately following the ratio of data budgets (50K/10K). \textit{{Table~\ref{tab:embedding_ablation}} further shows that using \textbf{the \textit{Select} stage alone} reduces gFID from 37.07 to 14.96, underscoring its effectiveness and usefulness.}

\noindent\textbf{Ablation on \textit{Attach} Phase.} We evaluate \textit{Attach} from two angles. First, as shown in Fig.~\ref{fig:interval&componets_ablation} (Right), DC embedding consistently outperforms using either alone under a 10K budget, with text-only better than class-only, indicating richer semantics from textual descriptions. Second, Table~\ref{tab:embedding_ablation} shows steady gains from the injection modules: baseline gFID-10K is 14.96, adding only visual information reaches 10.37, adding only DC embedding reaches 9.01, and combining both achieves the best 7.62. Appendix~\ref{appendix:visual information} further ablates the visual encoder and demonstrates the robustness of our approach.

\begin{table}[t]
    \centering
    \scriptsize   
    \caption{Ablation studies on the Select and Attach phases. Sel.: Select. Vis.: Vision.}
    \label{tab:embedding_ablation}
    \vspace{-2pt}
    \setlength{\tabcolsep}{1pt}
    \resizebox{0.75\linewidth}{!}{%
        \begin{tabular}{lcccc}
        \toprule
        Model & Sel. & DC Emb. & Vis. Emb. & gFID$\downarrow$ \\
        \midrule
        DiT-L/2 & \XSolidBrush  & \XSolidBrush      & \XSolidBrush      & 37.07 \\
        DiT-L/2 & \XSolidBrush      & \Checkmark & \Checkmark  & 8.79 \\
        \midrule
        DiT-L/2 & \Checkmark & \XSolidBrush      & \XSolidBrush      & 14.96 \\
        DiT-L/2 & \Checkmark  & \XSolidBrush      & \Checkmark  & 10.37 \\
        DiT-L/2 & \Checkmark & \Checkmark & \XSolidBrush      & 9.01 \\
        \CC DiT-L/2 &\CC \Checkmark & \CC\Checkmark &\CC \Checkmark &\CC \textbf{7.62} \\
        \bottomrule
        \end{tabular}
    }%
    \vspace{-8pt}
\end{table}

\begin{table}[t]
    \centering
    \small
    \caption{A breakdown of the computational overhead for sub-processes in D$^2$C. Compared to the REPA baseline, the additional scoring time is negligible, demonstrating D$^2$C's efficiency.}
    \label{tab:time}
    \vspace{-5pt}
    \setlength{\tabcolsep}{1pt}
    \resizebox{\linewidth}{!}{%
        \begin{tabular}{lccccc}
        \toprule
        Method & Score Model & Score Time & Iter. & Train Time & gFID$\downarrow$ \\
        \midrule
        REPA            & N/A           & N/A  & 4M     & 750h & 5.9 \\
        \makecell[l]{D\textsuperscript{2}C\\(w/o select)} & N/A           & N/A  & \textbf{0.04M}  & \textbf{7.4h} & 5.6 \\
        \makecell[l]{D\textsuperscript{2}C\\(w/ select)}          & From Scratch  & \textbf{1.9h} & \textbf{0.04M}  & \textbf{(7.4+26.2)h} & 4.9 \\
        \CC \makecell[l]{D\textsuperscript{2}C\\(w/ select)} & \CC Pretrained & \CC 2.1h & \CC \textbf{0.04M} & \CC \textbf{7.4h} & \CC \textbf{4.3}\\
        \bottomrule
        \end{tabular}
    }%
    \vspace{-15pt}
\end{table}

\noindent{\bf Effect of Pretrained Diffusion Models and Wall-Clock Cost.} Our D$^2$C pipeline does not inherently require a powerful pretrained model. As shown in Table~\ref{tab:time}, when the scoring network is a strong DiT-XL/2 with base gFID 2.27 from~\citep{DIT}, D$^2$C reaches an FID of 4.3; with a weaker DiT-L/2 that we trained from scratch achieving a base gFID of 11.5, it reaches 4.9. Using only the \textit{Attach} stage, without \textit{Select}, still reaches 5.6 and surpasses REPA at 5.9. In wall-clock terms, the \textit{Attach}-only variant finishes in 7.4h, which is 0.99\% of REPA’s 750h and about 101$\times$ faster. With a pretrained scorer, the end-to-end pipeline totals 9.5h, with 2.1h for scoring and 7.4h for training; this is 1.27\% of REPA and about 79$\times$ faster. With a scorer trained from scratch, the pipeline totals 35.5h, with 1.9h for scoring, 26.2h for training the scorer, and 7.4h for diffusion training; this is 4.7\% of REPA and about 21$\times$ faster. These results show that whether the scorer is strong, weak, or omitted, D$^2$C consistently accelerates diffusion training while maintaining competitive quality.

%% file: sec/5_conclusion.tex
\section{Conclusion}
\label{sec:conclusion}

In this paper, we introduce \textit{D$^2$C}, the first dataset condensation framework that significantly accelerates diffusion model training for generative tasks. \textit{D$^2$C} follows a two-phase pipeline, \textit{Select} and \textit{Attach}, which selects a compact yet diverse subset via a diffusion difficulty score with interval sampling and enriches it with semantic and visual signals. On ImageNet-1K, \textit{D$^2$C} achieves $100\text{--}233\times$ faster training than strong baselines while maintaining competitive generative quality, and we hope it will motivate further research on data-centric efficiency for diffusion models.

\textbf{Acknowledgement.} This work was supported by the National Natural Science Foundation of China under Grant No. 62506317.
 

%% file: sec/X_suppl.tex
\clearpage
\setcounter{page}{1}
\appendix
\maketitlesupplementary

% \section{Limitation}
% \label{sec:limitation}

% This work primarily focuses on accelerating and improving the training process of diffusion models. While our approach demonstrates promising results in the early stage, the final performance upon convergence does not yet fully match baseline benchmarks. It is important to unleash the potential of our algorithm, such as fine-tuning the vision and text encoders to better align noise latent features with the integrated visual information, thereby reducing existing discrepancies. Additionally, although the current study centers on image generation tasks, extending our method to other domains, such as 3D generation and video synthesis, represents a promising direction for future exploration.

\section{Positioning D$^2$C within Dataset Condensation Paradigms}

While some works equate dataset condensation with gradient-based pixel-level optimization of synthetic images, a broader line of literature defines it as constructing compact training sets that retain the learning efficacy of the original data~\citep{wu2025dc3}, which also includes image-level schemes such as OD3\citep{OD3} and RDED\citep{RDED}. In this broader paradigm, the key objective is not how the condensed data are obtained, but whether the resulting small dataset can support training models that closely match the performance of those trained on the full dataset. D$^2$C follows this latter view. It condenses the dataset by selecting a highly informative subset guided by diffusion difficulty and then attaching additional semantic and visual information that enriches each sample without altering its raw pixels. This design is analogous in spirit to OD3 and RDED, which also operate at the level of image selection rather than direct pixel optimization. Consequently, D$^2$C naturally fits within the dataset condensation family, while being specifically tailored to generative diffusion models and addressing a gap that is not covered by existing pixel-level condensation methods.

\section{Additional Descriptions of Diffusion Models}\label{AppendixA}
This section reviews the fundamentals of the Denoising Diffusion Probabilistic Model (DDPM)~\citep{ddpm_begin}. The DDPM framework consists of a fixed forward process that incrementally perturbs the input data with noise, and a learned reverse process trained to iteratively denoise the data, thereby learning the target distribution. Specific architectural details of our implementation are summarized in Appendix~\ref{appendix:architecture_detail}.

\subsection{Denoising Diffusion Probabilistic Model}

The DDPM framework models data generation via a discrete-time Markov chain that progressively adds Gaussian noise to a data sample $x_0 \sim p(x)$. The forward process is defined as:
\begin{equation}
q(x_t \mid x_{t-1}) = \mathcal{N}(x_t; \sqrt{1 - \beta_t} x_{t-1}, \beta_t \mathbf{I}),
\end{equation}

where $\beta_t \in (0, 1)$ are predefined variance schedule parameters controlling the noise level at each time step $t \in [1, 2, ..., T]$, and $\mathbf{I}$ is the identity matrix.

For simplicity, we define $\alpha_t = 1 - \beta_t$, and denote the cumulative product $\bar{\alpha}_t = \prod_{i=1}^t \alpha_i$. The reverse process, which is learned by the model $\theta$, can be defined as:

\begin{equation}
\resizebox{.95\linewidth}{!}{$
    p_\theta(x_{t-1} \mid x_t) = \mathcal{N} \left( x_{t-1}; \frac{1}{\sqrt{\alpha_t}} \left( x_t - \frac{\beta_t}{\sqrt{1 - \bar{\alpha}_t}} \epsilon_\theta(x_t, t) \right), \Sigma_\theta(x_t, t) \right),
$}
\end{equation}

where $\epsilon_\theta(x_t, t)$ denotes the predicted noise from a neural network. The covariance $\Sigma_\theta(x_t, t)$ is typically set to $\sigma^2_t\mathbf{I}$, where $\sigma^2_t$ can be either fixed~($\sigma^2_t = \beta_t$) or learned through interpolation $\sigma^2_t = (1 - \bar{\alpha}_{t-1}) / (1 - \bar{\alpha}_t)\beta$.

A simplified training objective minimizes the prediction error between true and estimated noise:
\begin{equation}
% \mathcal{L}_{\text{simple}} = \mathbb{E}_{x_0, \epsilon, t} \left[ \left\| \epsilon - \epsilon_\theta(x_t, t) \right\|^2 \right].  
\mathcal{L}_{\text{simple}} = \mathbb{E}_{x_0, \epsilon, t} \left[ \Vert \epsilon - \epsilon_\theta\left( \sqrt{\bar{\alpha}_t}x_0 + \sqrt{1-\bar{\alpha}_t}\epsilon, t \right) \Vert^2 \right].
\end{equation}

In addition to the simple objective, improved variants include learning the reverse variance $\Sigma_\theta(x_t, t)$ jointly with the mean, which leads to a variational bound loss of the form:
\begin{equation}
\mathcal{L}_{\text{vlb}} = \exp\left( v \log \beta_t + (1 - v) \log \tilde{\beta}_t \right).
\end{equation}

Here, $v$ is an element-wise weight across model output dimensions. When $T$ is sufficiently large and the noise schedule is carefully chosen, the terminal distribution $p(x_T)$ approximates an isotropic Gaussian. Sampling is then performed by iteratively applying the learned reverse process to recover the data sample from pure noise.

\subsection{Diffusion Transformer Architecture}
\label{appendix:architecture_detail}

Our model implementation closely follows the design of DiT~\citep{DIT} and SiT~\citep{SiT}, which extend the vision transformer (ViT) architecture~\citep{VIT} to generative modeling. An input image is first split into patches, reshaped into a 1D sequence of length $N$, and then processed through transformer layers. To reduce spatial resolution and computational cost, we follow prior work~\citep{DIT,SiT} and encode the image into a latent tensor $z = E(x)$ using a pretrained encoder $E$ from the stable diffusion VAE.

In contrast to the standard ViT, our transformer blocks include time-aware adaptive normalization layers known as adaLN-zero. These layers scale and shift the hidden state in each attention block according to the diffusion timestep and conditioning signals. During training, we also add an auxiliary multilayer perceptron (MLP) head that maps the hidden state to a semantic target representation space, such as DINOv2~\citep{dinov2} or CLIP features~\citep{CLIP}. This head is used only for training-time supervision in our alignment loss and does not affect sampling or inference.

\section{Hyperparameters and Implementation Details}
\label{Implmentation_details}

\noindent{\textbf{Select Phase Settings.}} In the \textit{Select} phase, we adopt a pre-trained DiT-XL/2 model~\citep{DIT} as the scoring network and use the diffusion loss (\textit{w.r.t.}, mean squared error) as the scoring metric. To construct subsets of different sizes, we apply interval sampling with $k = 96$ for the 10K subset, $k = 16$ for the 50K subset, and $k = 10$ for the 100K subset. Each subset is constructed in a class-wise manner, selecting 10, 50, and 100 samples per class respectively.

\noindent{\textbf{Attach Phase Settings.}} In the \textit{Attach} phase, we implement dual conditional embeddings. For textual conditioning, we use a T5 encoder~\citep{ni2021sentence} with captions truncated to 16 tokens, producing embeddings of dimension 2048. For visual conditioning, we adopt DINOv2-B~\citep{dinov2} as the visual encoder. The number of visual tokens $h$ is set to 256, and each token has a feature dimension of 768.

\noindent{\textbf{Training Settings.}} In the \textit{Training} phase, we use the Adam optimizer with a fixed learning rate of 1e-4 and $(\beta_1, \beta_2) = (0.9, 0.999)$, without applying weight decay. We employ mixed-precision (fp16) training with gradient clipping. Latent representations are pre-computed using the stable diffusion VAE, and decoded via its native decoder. All experiments are conducted on either 8 NVIDIA A800 80GB GPUs or 8 NVIDIA RTX 4090 24GB GPUs. We use a batch size of 256 with a $256 \times 256$ resolution in Fig.~\ref{fig:first}, and a $512 \times 512$ resolution in Table~\ref{tab:512Xcomparision}. All other experiments use a batch size of 128 and a default image resolution of $256 \times 256$.

\section{Evaluation Details}
\label{evaluation_details}

We adopt several widely used metrics to evaluate generation quality and diversity:
\begin{itemize}[leftmargin=1.5em, itemsep=2pt, topsep=2pt]
    \item \textbf{gFID}~\citep{fid} computes the Fréchet distance between the feature distributions of real and generated images. Features are extracted using the Inception-v3 network~\citep{googlenetv2}.
    \item \textbf{sFID}~\citep{sFiD} extends FID by leveraging intermediate spatial features from the Inception-v3 model to better capture spatial structure and style in generated images.
    \item \textbf{IS}~\citep{is} evaluates both the quality and diversity of generated samples by computing the KL-divergence between the conditional label distribution and the marginal distribution over predicted classes, using softmax-normalized logits.
    \item \textbf{Precision and Recall}~\citep{pre} respectively measure sample realism and diversity, quantifying how well generated samples cover the data manifold and vice versa.
\end{itemize}

\section{Baseline Setting}
\label{Baeline_details}
We evaluate our method against two categories of baselines:

\noindent{\textbf{Diffusion models trained on selected or condensed subsets.}} These include SiT and DiT backbones trained from scratch on 10K, 50K, and 100K subsets obtained via the following strategies:
\begin{itemize}[leftmargin =1.5em, itemsep=2pt, topsep=2pt]
    \item \textbf{Random Sampling.} A naive baseline that randomly selects a fixed number of real samples without any guidance.

    \item \textbf{Herding}~\citep{herding}. A geometry-based method that selects samples to approximate the global feature mean, ensuring representative coverage.

    \item \textbf{K-Center}~\citep{kcenter}. A diversity-focused algorithm that iteratively selects samples maximizing the minimum distance from the selected set, promoting broad coverage of the feature space.

    \item \textbf{SRe\textsuperscript{2}L}~\citep{dd_sre2l}. A dataset condensation method that synthesizes class-conditional data through a multi-stage pipeline. Originally proposed for classification tasks, we adapt it to the diffusion setting by applying class-wise condensation to real images and training a diffusion model on the resulting synthetic subset. Visualizations of the synthesized samples and corresponding training results are provided in Appendix~\ref{appdix:concendation}.
\end{itemize}

\noindent{{\textbf{Diffusion models trained on the full dataset.} }}These baselines are trained with access to the entire training set, without data reduction:
\begin{itemize}[leftmargin=1.5em, itemsep=2pt, topsep=2pt]
    \item \textbf{SiT}~\citep{SiT}. A transformer-based diffusion model that reformulates denoising as continuous stochastic interpolation, enabling faster training and improved efficiency under full-data settings.
    
    \item \textbf{REPA}~\citep{REPA}. A model-side regularization method that aligns intermediate features of diffusion transformers with patch-wise representations from strong pretrained visual encoders (e.g., DINOv2-B~\citep{dinov2}, MAE~\citep{MAE}, MoCov3~\citep{MOCO}) using a contrastive loss. It retains the full dataset and improves convergence and generation quality via early-layer representation guidance.
\end{itemize}

\section{Framework Design and Implementation}
\label{appendix:theoretical}

We introduce D$^2$C, a framework for constructing compact yet effective training subsets for diffusion models under stringent data budgets. Our approach is motivated by two complementary intuitions: (1) that the contribution of training samples is non-uniform, as some are more informative than others; and (2) that generative training benefits from semantically enriched conditioning. These insights directly inform the two core stages of our framework. First, a \textit{Select} stage ranks training examples by a difficulty score computed via a pretrained class-conditional diffusion model. Second, an \textit{Attach} stage enriches the selected data by injecting textual and visual priors. The complete pipeline is summarized in Algorithm~\ref{alg:d2c}.

\begin{algorithm}[h]
\caption{D$^2$C: Diffusion Dataset Condensation}
\label{alg:d2c}
\begin{algorithmic}[1]
\REQUIRE Full dataset $\mathcal{D} = \{(x_i, c_i)\}_{i=1}^N$, interval $k$, text encoder $f_\text{text}$, visual encoder $f_\text{vis}$ \\
\textit{// Each $x_i$ is an image, and $c_i \in \{1, \dots, C\}$ is the class label.}

\vspace{2pt}
\STATE {\bf // Phase 1: Select}
\STATE Compute difficulty score $s_\text{diff}$ for all $(x_i, c_i) \in \mathcal{D}$
\STATE For each class $c$, sort $\mathcal{D}_c = \{x_i \mid c_i = c\}$ by $s_\text{diff}$ ascending
\STATE Select every $k$-th sample (Interval Sampling) in sorted $\mathcal{D}_c$ to form $\mathcal{D}_\text{select}$

\vspace{2pt}
\STATE {\bf // Phase 2: Attach}
\FOR{each $(x, c) \in \mathcal{D}_\text{select}$}
    \STATE Generate class prompt $P(c)$ (e.g., ``a photo of a \texttt{label}'')
    \STATE Extract text embedding: $(t_c, t_\textrm{mask}) \gets f_\text{text}(P(c))$
    \STATE Extract visual feature: $y_\text{vis} \gets f_\text{vis}(x)$
    \STATE Store triplet $(x, c, t_c, t_\textrm{mask}, y_\text{vis})$ into $\widetilde{\mathcal{D}}$
\ENDFOR

\vspace{2pt}
\STATE {\bf Return} enriched dataset $\widetilde{\mathcal{D}}$ for diffusion model training
\end{algorithmic}
\end{algorithm}

\section{Exploration on Text-to-Image Generation}
\label{sec:select_t2i}

\begin{figure}[t]
  \centering
  \includegraphics[width=\linewidth]{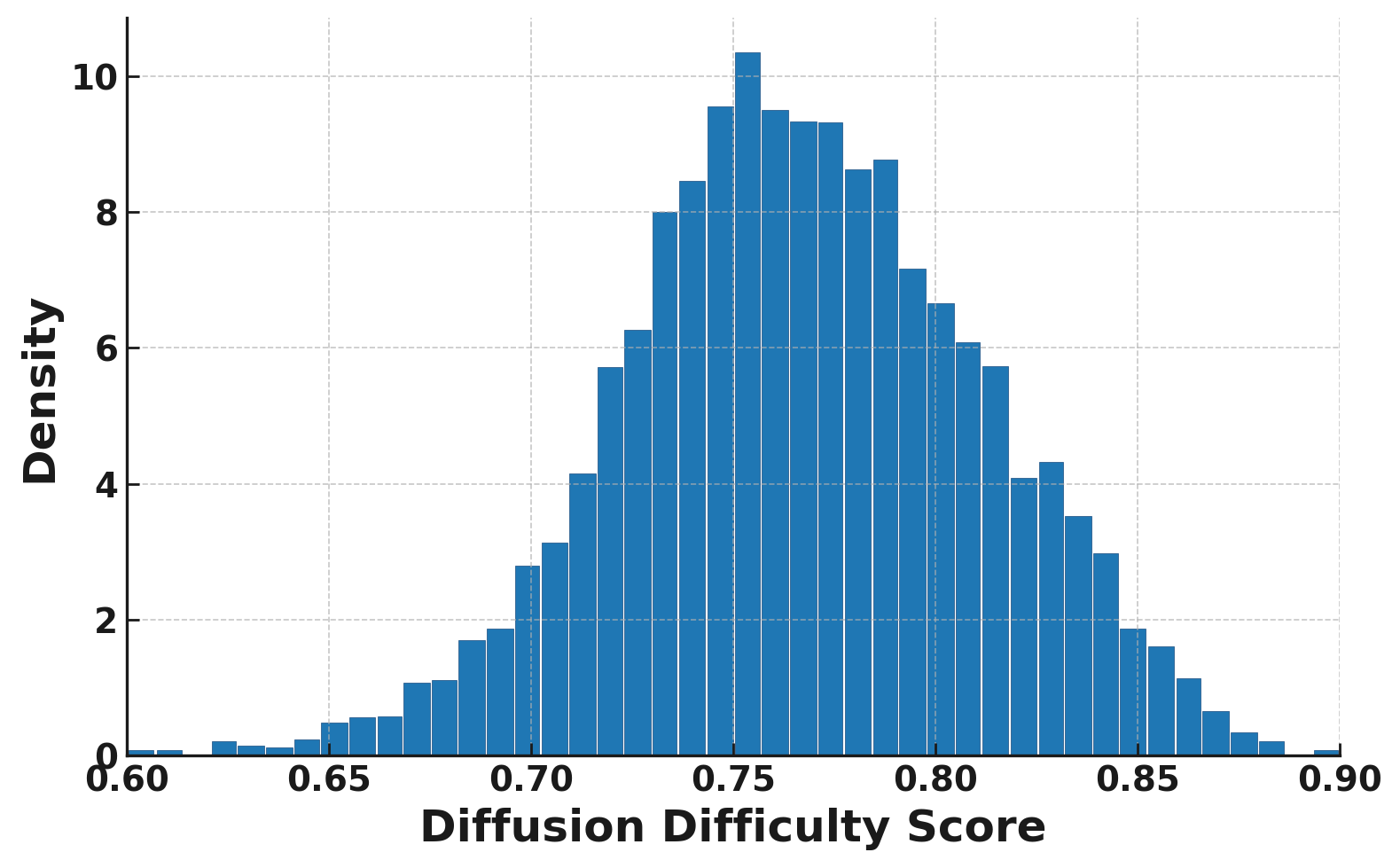}
  \vspace{-8pt}
    \caption{Distribution of diffusion difficulty score computed on LAION text–image pairs with a pre-trained SDXL model. This distribution resembles that of C2I, which supports interval sampling for selecting informative training pairs under T2I.}
  \label{fig:t2i}
  \vspace{-5pt}
\end{figure}

{We further examine the applicability of the \textbf{\textit{D\textsuperscript{2}C}} framework to text-to-image generation~\citep{xie2026guidance,shi2025culture, fang2025flexcontrol}. The \textit{Select} phase requires only a minimal change: replace the class condition in Eq.~\ref{eq-sdiff} with a text condition, i.e.,
\( s_{\mathrm{diff}}^{text}(x) = -p_{\theta}\!\left(x \mid \text{text}\right) \).
Using SDXL to score LAION text--image pairs, we observe a difficulty distribution similar to the class-conditional case (Fig.~\ref{fig:t2i}; see also Fig.~\ref{fig:loss_vis} and Fig.~\ref{fig:interval_ablation}~(right)). Low-score samples tend to exhibit simple structures, high-score samples often contain complex or cluttered contexts, and the majority of samples fall in the middle range. Interval sampling remains effective for identifying informative pairs. The \textit{Attach} phase is also easy to transfer: semantic and visual representations serve as soft supervisory signals for the selected subset.}

{As such, while our main experiments focus on class-to-image tasks for controlled benchmarking like SiT~\citep{SiT}, the framework is generalizable and well suited to text-to-image generation. We expect it to deliver practical gains in data efficiency and training speed in this setting, offering a promising direction for future work.}

\section{More Discussions about Select}

\subsection{{Detailed Algorithm for Computing Diffusion Difficulty Score}}
\label{app:dds_algo}

The diffusion difficulty score, used to rank samples in the \textit{Select} phase, is defined as the mean denoising loss over uniformly sampled timesteps, computed using a frozen pretrained diffusion model (see Algorithm~\ref{alg:dds}).

\begin{algorithm}[h]
\caption{{Compute Diffusion Difficulty Score}}
\label{alg:dds}
\begin{algorithmic}[1]
\REQUIRE Image dataset $\mathcal{D}=\{(x_i,c_i)\}_{i=1}^N$; pretrained VAE encoder $E_\phi$; pretrained diffusion model $\epsilon_\theta$; timestep set $\mathcal{T}$; batch size $n$ \\
\textit{// Each $x_i$ is an image; $c_i \in \{1,\dots,C\}$ is the class label. Timesteps in $\mathcal{T}$ are sampled uniformly. Models are frozen during scoring.}

\vspace{2pt}
\STATE Initialize empty map $\mathcal{S} \leftarrow \{\}$

\vspace{2pt}
\FOR{mini-batch $\{(x_i,c_i)\}_{i=1}^n \subset \mathcal{D}$}
    \STATE Encode to latent (if applicable): $z_i \gets E_\phi(x_i)$ \ 
    \STATE Initialize per-sample accumulator $\ell_i \gets 0$
    \FOR{$t \in \mathcal{T}$}
        \STATE Sample $\epsilon \sim \mathcal{N}(0,I)$
        \STATE Perturb latent: $z_t \gets \alpha_t \, z_i + \sigma_t \, \epsilon$
        \STATE Compute loss: $\ell_i \gets \ell_i + \lVert \epsilon - \epsilon_\theta(z_t, t, c_i) \rVert_2^2$
    \ENDFOR
    \STATE $s_i \gets \ell_i / |\mathcal{T}|$ \ \textit{// Mean denoising loss across timesteps}
    \STATE $\mathcal{S}[x_i] \gets s_i$
\ENDFOR

\vspace{2pt}
\STATE \textbf{Return} $\mathcal{S}$ \ \textit{// Image-to-score mapping for difficulty-aware selection}
\end{algorithmic}
\end{algorithm}

\subsection{Practical Insights on Interval Sampling}
\label{app:select_insights}

\begin{figure}[h]
  \centering
  \includegraphics[width=1.0\linewidth]{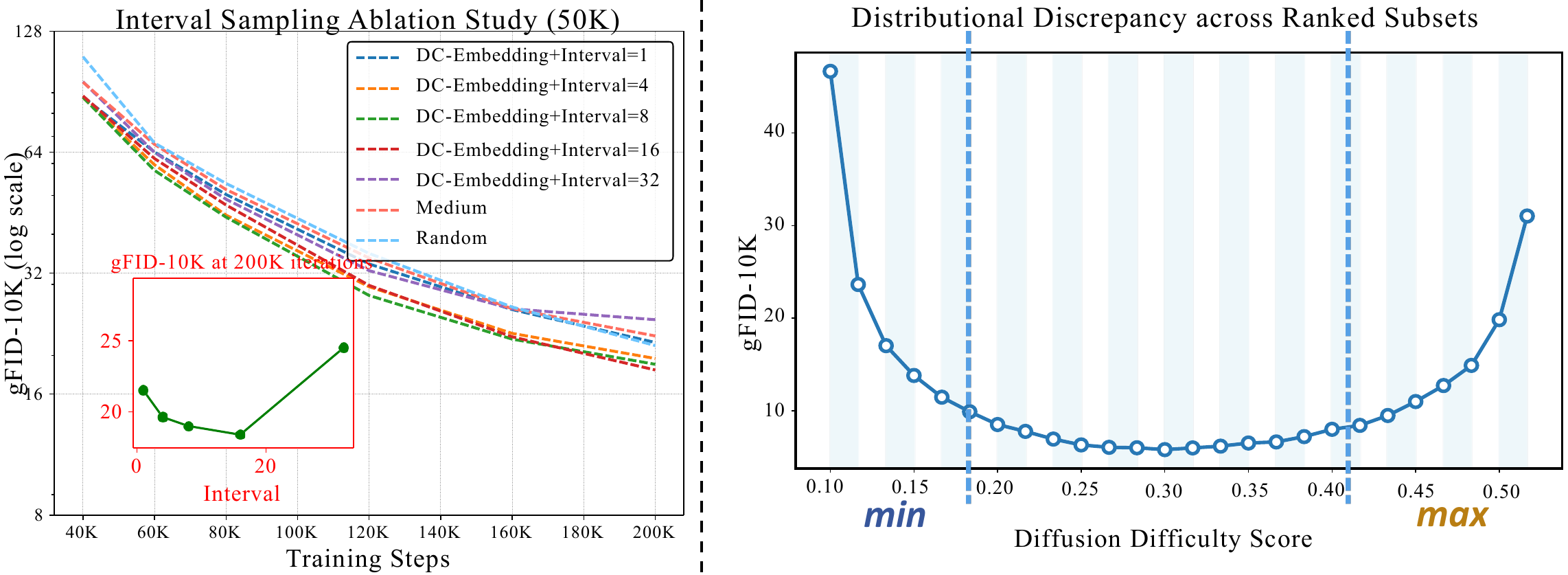}
  \vspace{-14pt}
    \caption{\textbf{Left:} gFID-10K across training steps under different interval values $k$ for a 50K data budget. Moderate intervals (e.g., $k=16$) achieve superior performance by balancing learnability and diversity. \textbf{Right:} Distributional discrepancy (gFID-10K) between ranked training subsets and the validation set. Both extremely low and high diffusion difficulty score lead to higher FID, while mid-range segments show better alignment.}
  \label{fig:interval_ablation}
  \vspace{-5pt}
\end{figure}

While Section~\ref{sec:ablation} has covered a detailed ablation study on the choice of interval $k$ in \textit{Select} phase, we provide additional insights into how diffusion difficulty score relates to distributional coverage.

The right panel in Fig.~\ref{fig:interval_ablation} presents the gFID-10K scores of subsets sampled from different portions of the difficulty-ranked dataset. We partition the training set into consecutive 10K segments ordered by the diffusion difficulty score (e.g., the first 10K samples with lowest scores as “Min”, followed by 10–20K, 20–30K, and so on), and measure each segment’s discrepancy from the full validation distribution using gFID. Interestingly, we observe a clear U-shaped curve: subsets consisting of extremely low or high difficulty samples exhibit significantly worse distributional alignment, while those centered around moderate difficulty levels show substantially lower FID scores. This result aligns well with our hypothesis that very easy samples (e.g., simple textures, clean backgrounds) and extremely hard samples (e.g., ambiguous, noisy structures) both fail to reflect the global data distribution.

These observations provide an empirical justification for our interval sampling strategy. Specifically, under a 50K dataset budget with $k = 16$, each class contributes samples selected at regular intervals from its difficulty-sorted list. Given that each class typically contains around 1,200 images, this strategy naturally samples from approximately the first 800 positions in the ranked list. As a result, the selected data span both the easy and moderately difficult regions, while avoiding the extremes at both ends. This balanced coverage across the difficulty spectrum promotes better generalization and faster convergence, as evidenced by the results in Fig.~\ref{fig:interval_ablation} (Left) and discussed in Section~\ref{sec:ablation}. In this way, our strategy yields a compact yet effective dataset that enables the model to converge rapidly while maintaining strong generation quality.

\textbf{Ablation on interval sampling.} As shown in Fig.~\ref{fig:tsne}, the ``Medium'' variant corresponds to selecting samples from the center of the difficulty-ranked list rather than applying interval sampling from low to high diffusion difficulty scores. Concretely, after sorting each class by diffusion difficulty, we start from the median position and expand symmetrically toward both sides until the data budget is reached. This strategy focuses on medium-difficulty examples and largely omits easier instances, while still including a portion of harder ones near the tails. As a result, the selected subset provides less comprehensive coverage of the underlying data distribution, leading to slower convergence and degraded final performance compared to our proposed interval sampling scheme.

\section{More Discussions about Attach}

\subsection{Dual Conditional Embedding}
\label{appendix:DC embedding}

\begin{figure}[t]
  \centering
  \includegraphics[width=\linewidth]{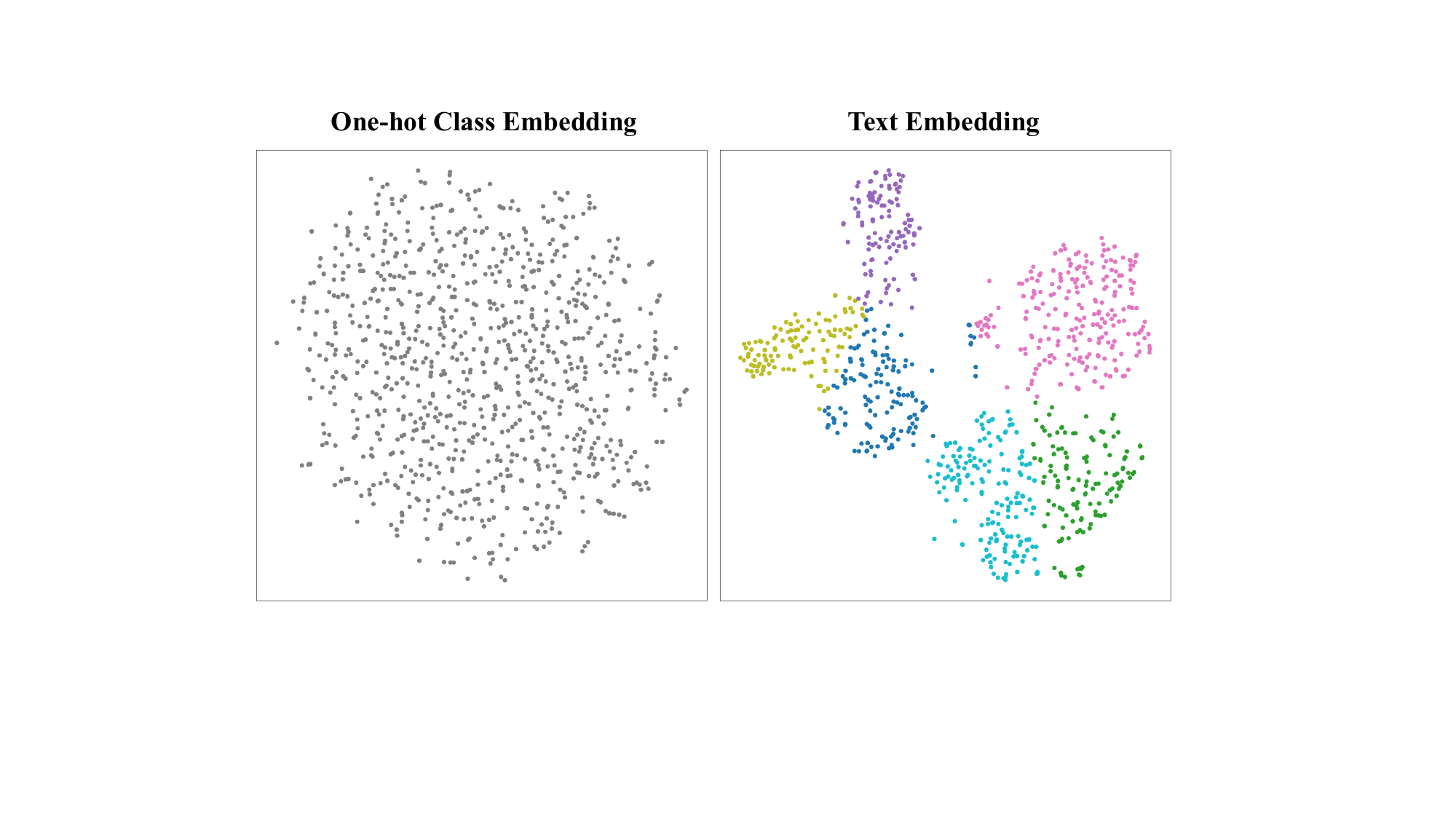}
  \vspace{-8pt}
  \caption{T-SNE visualization of class embeddings. Each point represents a class in the dataset. \textbf{Left:} One-hot class embeddings show no semantic structure. \textbf{Right:} Text embeddings naturally cluster semantically related classes. Samples from semantically related classes, such as different dog breeds, tend to form distinct clusters in feature space. Leveraging this semantic prior is highly effective for accelerating diffusion model training.}
  \label{fig:tsne}
  \vspace{-5pt}
\end{figure}

Most diffusion models condition on class identifiers represented as integer IDs or one-hot vectors, which are mapped to class embeddings trained from scratch. This ignores semantic relationships between categories, resulting in unstructured embeddings as shown in Fig.~\ref{fig:tsne} (Left).In contrast, text embeddings derived from class-specific prompts (e.g., “a photo of a dog”) via a pre-trained language encoder naturally encode semantic priors and cluster related classes (Fig.~\ref{fig:tsne}, Right). We propose a dual conditional embedding that fuses the text embedding with a learnable class embedding (i.e., a traditional class token trained from scratch),
 as defined in Eq.~\ref{eq:txet1}--\ref{eq:txet2}. This hybrid strategy combines semantic structure with symbolic distinctiveness, and leads to significantly improved generation quality. As shown in Fig.~\ref{fig:interval&componets_ablation} (Right), using both branches achieves lower FID than using either one alone.

\begin{figure*}[t]
  \centering
  \includegraphics[width=1.0\linewidth]{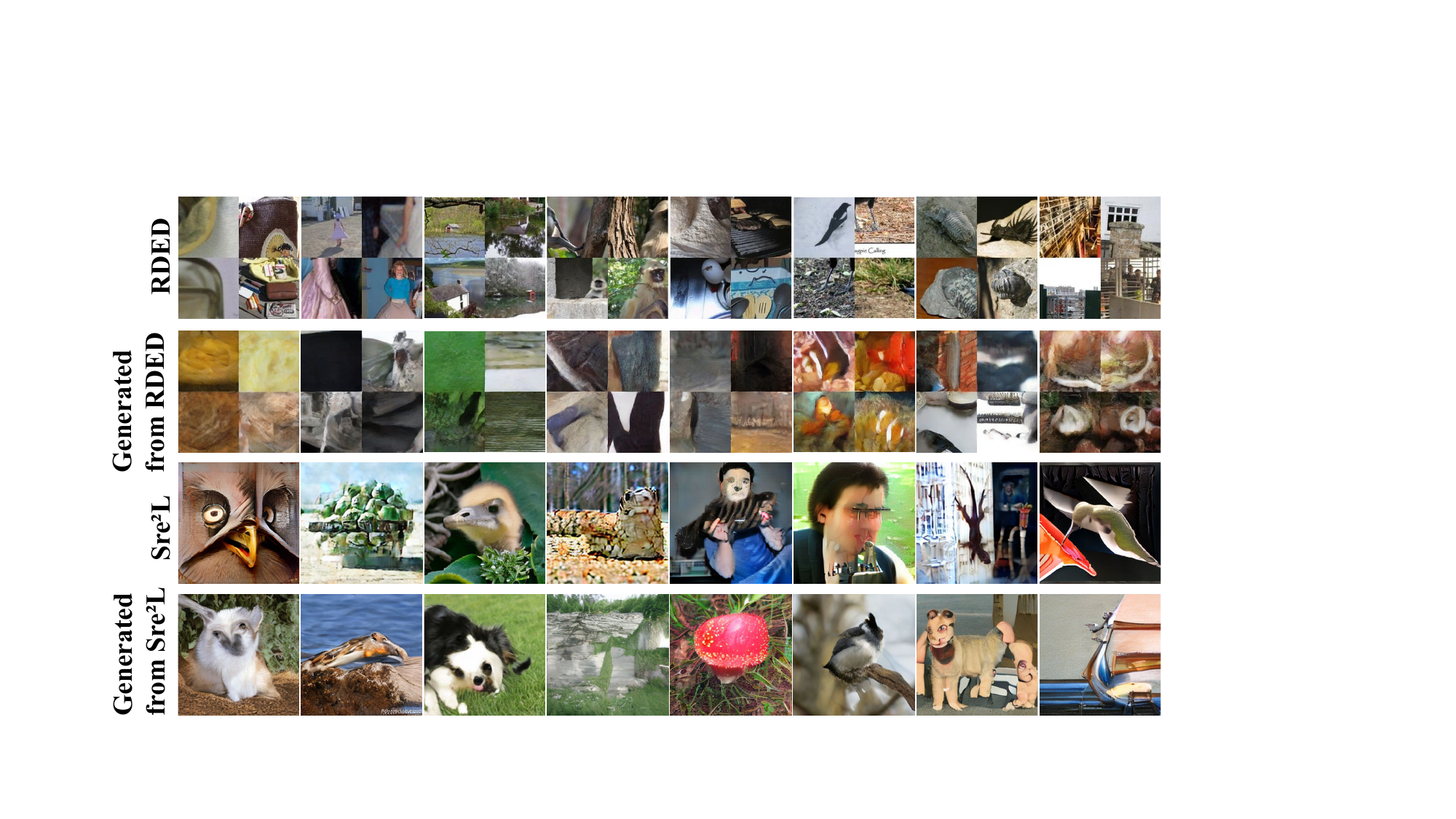}
  \vspace{-8pt}
    \caption{\textbf{Top:} Images synthesized directly by SRe$^2$L and RDED, two popular dataset condensation methods originally designed for discriminative tasks. \textbf{Bottom:} Images generated by  diffusion model trained on the two synthesized datasets.}
  \label{rded_sre2l}

  \vspace{-5pt}
\end{figure*}

\subsection{Visual Information Injection}
\label{appendix:visual information}

\begin{table}[htbp]
  \centering
  \caption{Ablation of the visual encoder.}
  \label{tab:vision-encoder}
  
  % 这里的 \small 已经去掉了，现在就是文档默认大小
  
  \begin{tabular}{l r}
    \toprule
    Vision Encoder & FID$\downarrow$ \\
    \midrule
    N/A (baseline) & 37.07 \\
    MAE-L          & 9.23  \\
    MoCov3-L       & 8.78  \\
    CLIP-L         & 8.59  \\
    \textbf{DINOv2-L} & \textbf{7.62} \\
    \bottomrule
  \end{tabular}
\end{table}

Recent studies~\citep{REG,REPA} have shown that relying solely on diffusion models to learn meaningful representations from scratch often results in suboptimal semantic features. In contrast, injecting high-quality visual priors, especially those derived from strong self-supervised encoders like DINOv2~\citep{dinov2}, can significantly improve both training efficiency and generation quality. In our case, we incorporate a frozen visual encoder to provide external patch-level visual features during training. These external features serve as semantically rich anchors, particularly beneficial at early layers, allowing the model to focus on generation-specific details in later stages. Empirically, visual supervision improves feature alignment and accelerates convergence under limited data, as shown in Tables~\ref{tab:ipc_comparison}, \ref{tab:512Xcomparision}, \ref{tab:embedding_ablation}, and \ref{tab:vision-encoder}. All tested encoders outperform the no-encoder baseline, indicating that our method is robust to the choice of visual encoder.

\section{Experiments on CIFAR}
\label{appdix:cifar}

As shown in Table~\ref{tab:cifar_d2c}, we further evaluate \textit{D\textsuperscript{2}C} on CIFAR-10 by selecting 100 images per class to form a 1K data budget (2\% compression rate) and training the diffusion model for 100k steps. Under this highly constrained setting, \textit{D\textsuperscript{2}C} significantly improves gFID from 9.72 with random sampling to 3.95, demonstrating that our selection and attachment strategy remains effective beyond ImageNet and transfers well across datasets.

\begin{table}[t]
\centering
\caption{Comparison of random subset selection and \textit{D$^2$C} on CIFAR-10 (reported in gFID-50K).}
\label{tab:cifar_d2c}
\begin{tabular}{lc}
\toprule
Method & gFID$\downarrow$ \\
\midrule
Random & 9.72 \\
\CC{\textit{D\textsuperscript{2}C} (Ours)} & \CC{\textbf{3.95}} \\
\bottomrule
\end{tabular}
\end{table}

\begin{figure*}[t]
  \centering
  \includegraphics[width=1.0\linewidth]{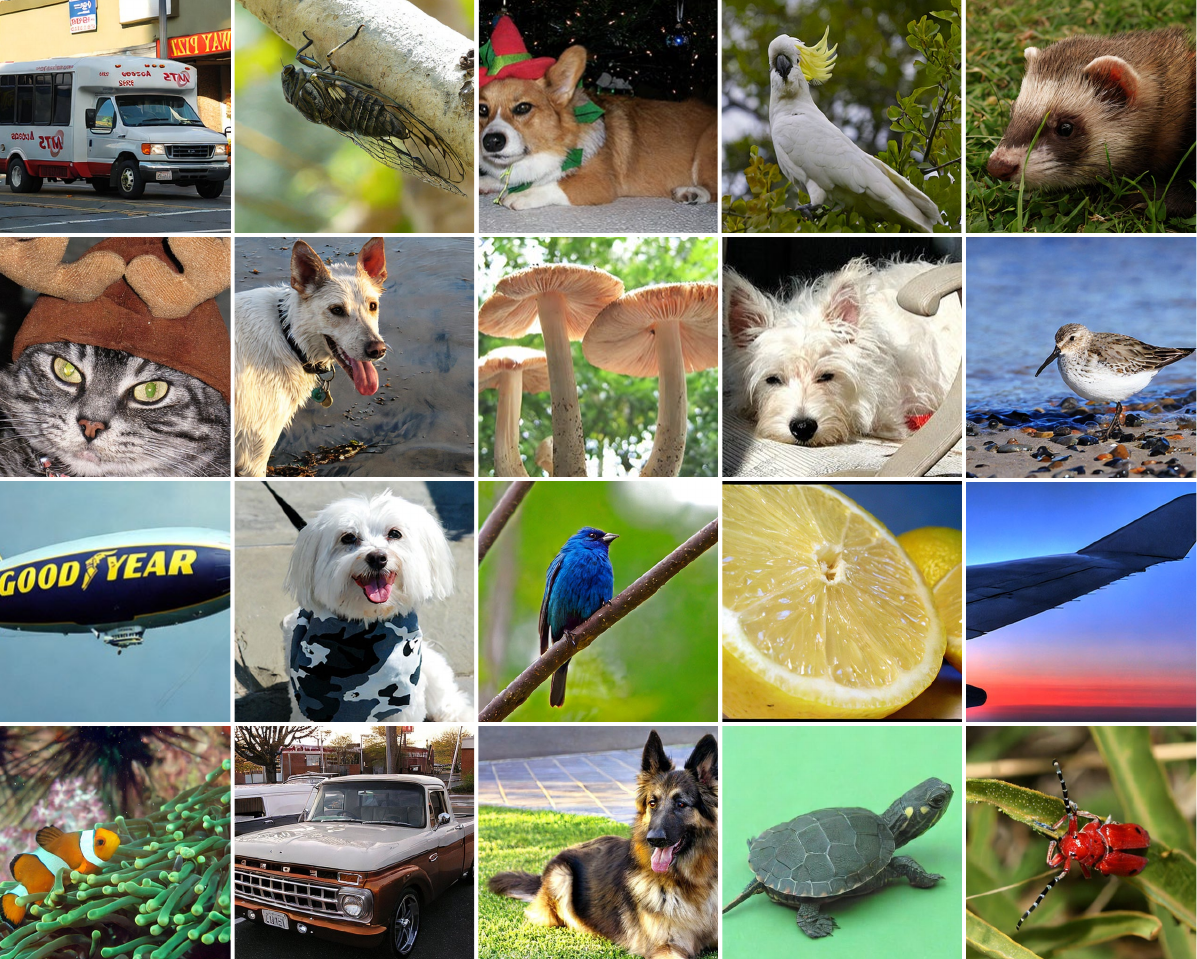}
  \vspace{-8pt}
    \caption{Generated samples on ImageNet 512$\times$512 from SiT-L/2 trained with \textit{D\textsuperscript{2}C} using a 10K dataset (CFG=1.5).}
    
  \label{fig:512}
  \vspace{-5pt}
\end{figure*}

\section{Visualization of SRe$^2$L and RDED in Generative Tasks}
\label{appdix:concendation}

As shown in Fig.~\ref{rded_sre2l}, dataset condensation methods that excel in classification, such as RDED and SRe$^2$L, transfer poorly to diffusion-based generation. Their objectives focus on preserving class-discriminative cues, for example segmentation-guided selection in RDED and gradient-based image optimization in SRe$^2$L, rather than modeling realistic global structure and natural image statistics. As a result, diffusion models trained on these synthesized datasets fail to capture the underlying pixel-level data distribution and produce severely degraded samples. In contrast, D$^2$C provides the first dataset condensation framework tailored to diffusion generative modeling and effectively closes this gap.

\section{\texorpdfstring{ImageNet 512$\times$512 Experiment}{ImageNet 512x512 Experiment}}
\label{appdix:imagenet_512}

As shown in Table~\ref{tab:512Xcomparision}, \textit{D\textsuperscript{2}C} consistently outperforms random sampling under a strict 10K (0.8\%) data budget across both DiT-L/2 and SiT-L/2 backbones. Visual samples in Fig.~\ref{fig:512} further confirm the high fidelity and diversity of generations at 512$\times$512 resolution, demonstrating that \textit{D\textsuperscript{2}C} generalizes effectively to high-resolution settings.

\section{Visualization}

\label{appdix:visualization}

\begin{figure*}[t]
  \centering
  \includegraphics[width=1.0\linewidth]{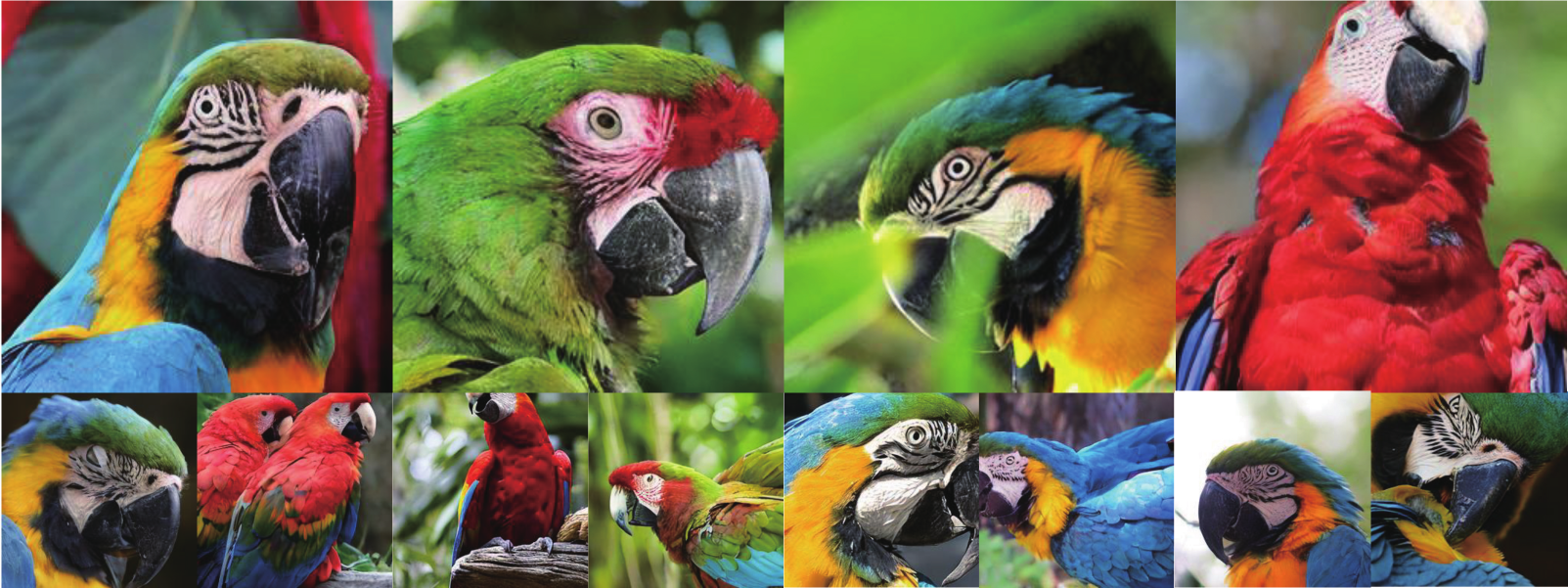}
  \vspace{-14pt}
    \caption{Generated samples of SiT-L/2 trained with \textit{D\textsuperscript{2}C} using a 50K dataset (CFG=1.5). Class label = "macaw"(88)}
  \vspace{-5pt}
\end{figure*}

\begin{figure*}[h]
  \centering
  \includegraphics[width=1.0\linewidth]{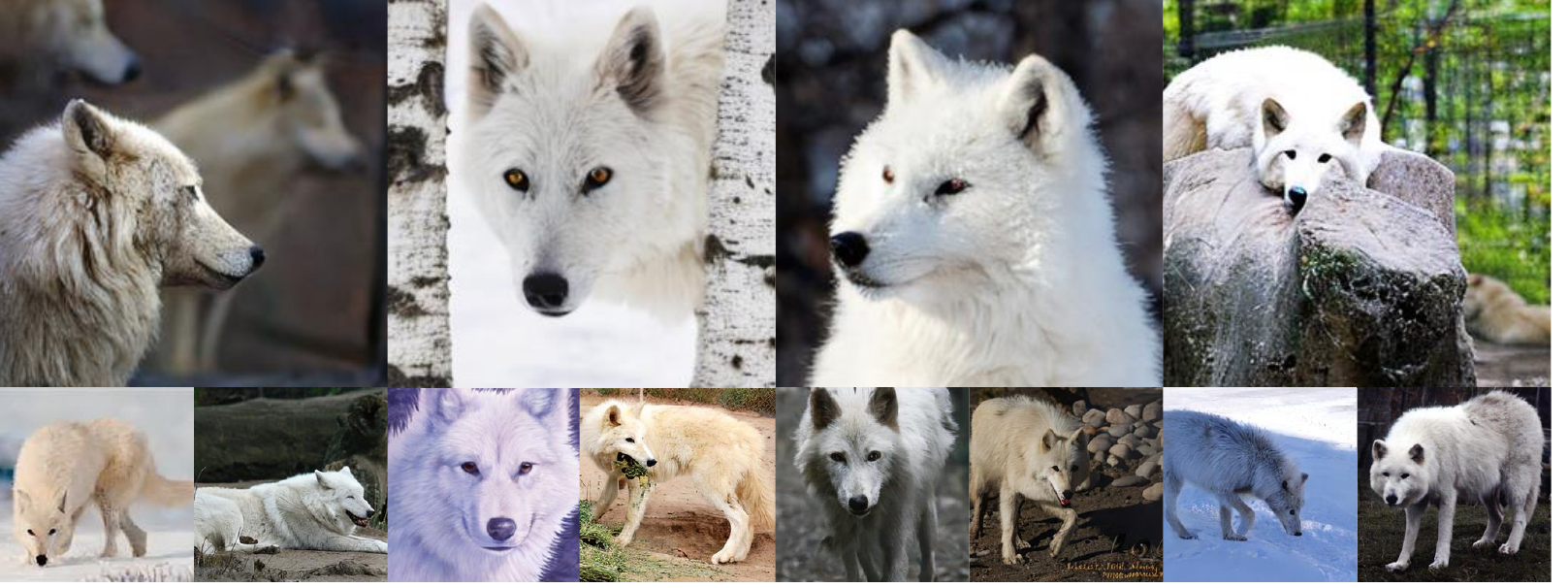}
  \vspace{-14pt}
    \caption{Generated samples of SiT-L/2 trained with \textit{D\textsuperscript{2}C} using a 50K dataset (CFG=1.5). Class label = "arctic wolf"(270)}
    
  \vspace{-5pt}
\end{figure*}

\begin{figure*}[t]
  \centering
  \includegraphics[width=1.0\linewidth]{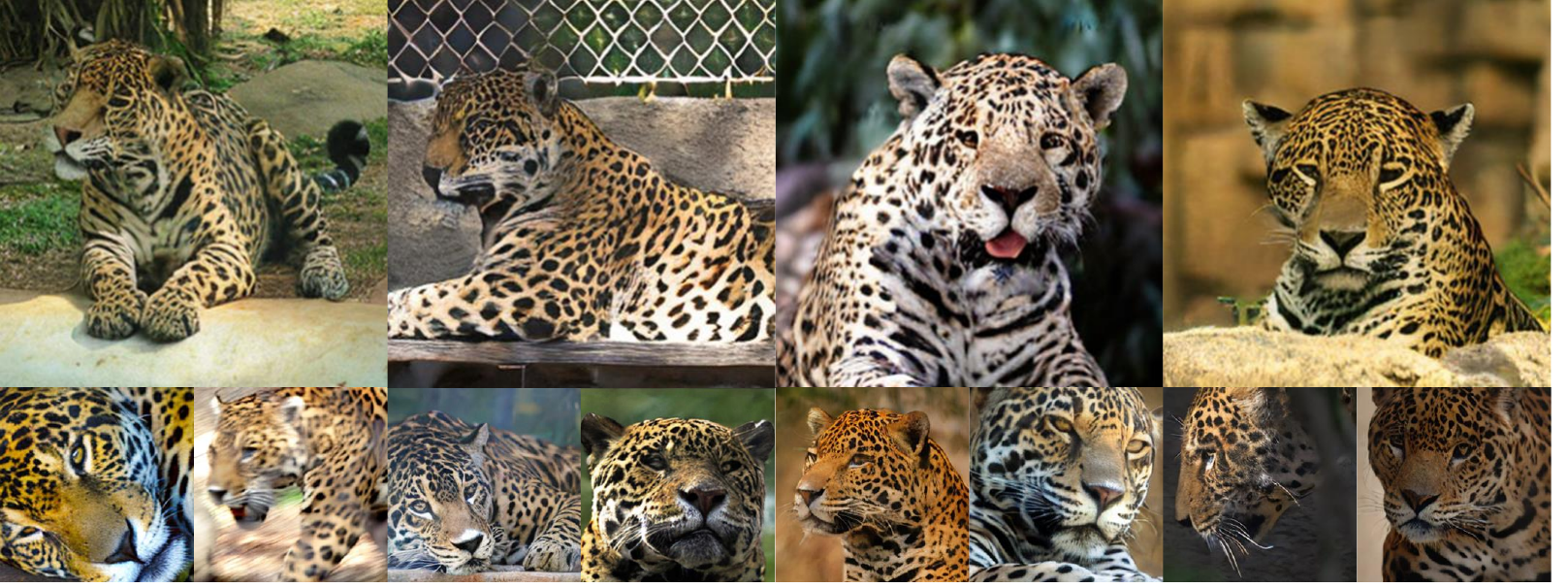}
  \vspace{-14pt}
    \caption{Generated samples of SiT-L/2 trained with \textit{D\textsuperscript{2}C} using a 50K dataset (CFG=1.5). Class label = "jaguar"(290)}
    
  \vspace{-5pt}
\end{figure*}

\begin{figure*}[t]
  \centering
  \includegraphics[width=1.0\linewidth]{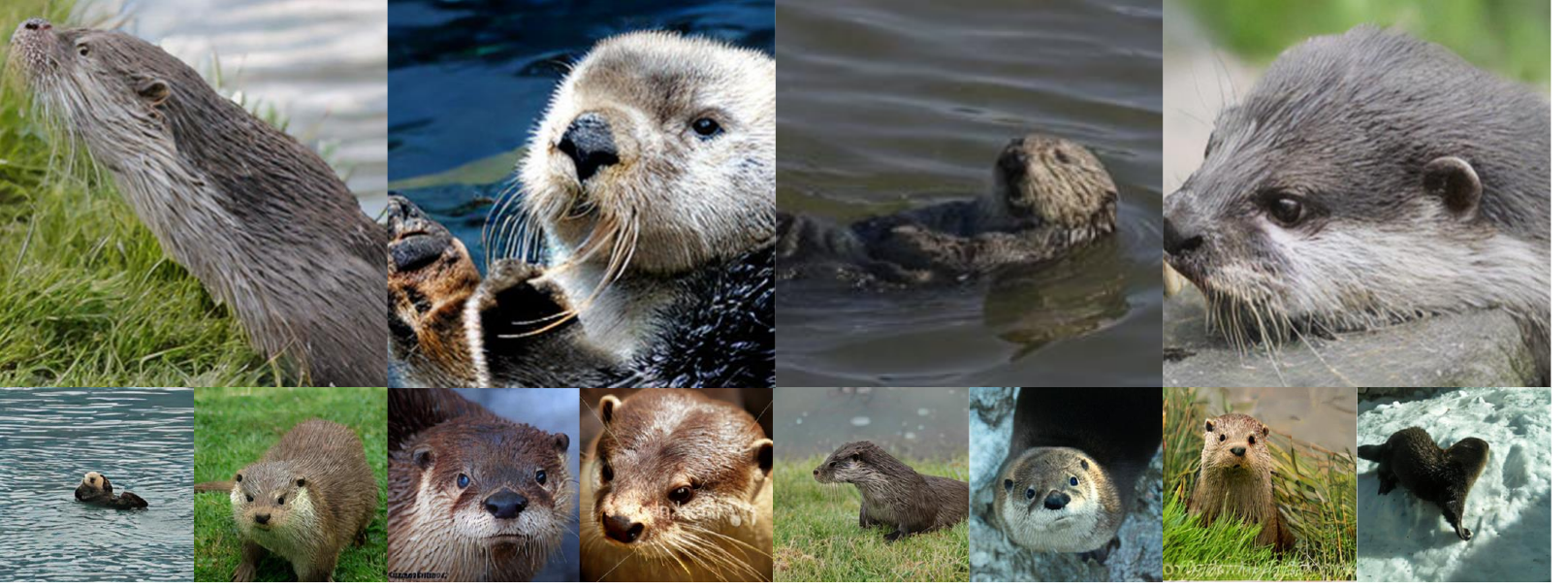}
  \vspace{-14pt}
    \caption{Generated samples of SiT-L/2 trained with \textit{D\textsuperscript{2}C} using a 50K dataset (CFG=1.5). Class label = "otter"(360)}
    
  \vspace{-5pt}
\end{figure*}

\begin{figure*}[t]
  \centering
  \includegraphics[width=1.0\linewidth]{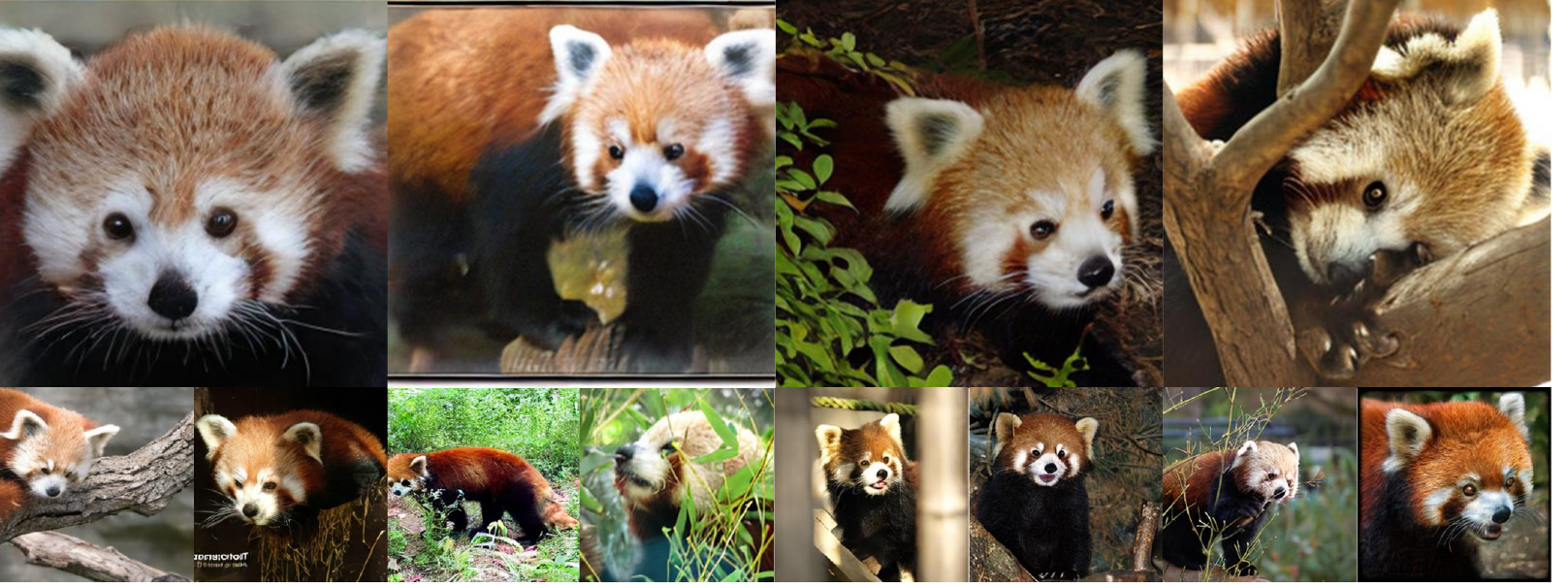}
  \vspace{-14pt}
    \caption{Generated samples of SiT-L/2 trained with \textit{D\textsuperscript{2}C} using a 50K dataset (CFG=1.5). Class label = "lesser panda"(387)}
    
  \vspace{-5pt}
\end{figure*}

\begin{figure*}[t]
  \centering
  \includegraphics[width=1.0\linewidth]{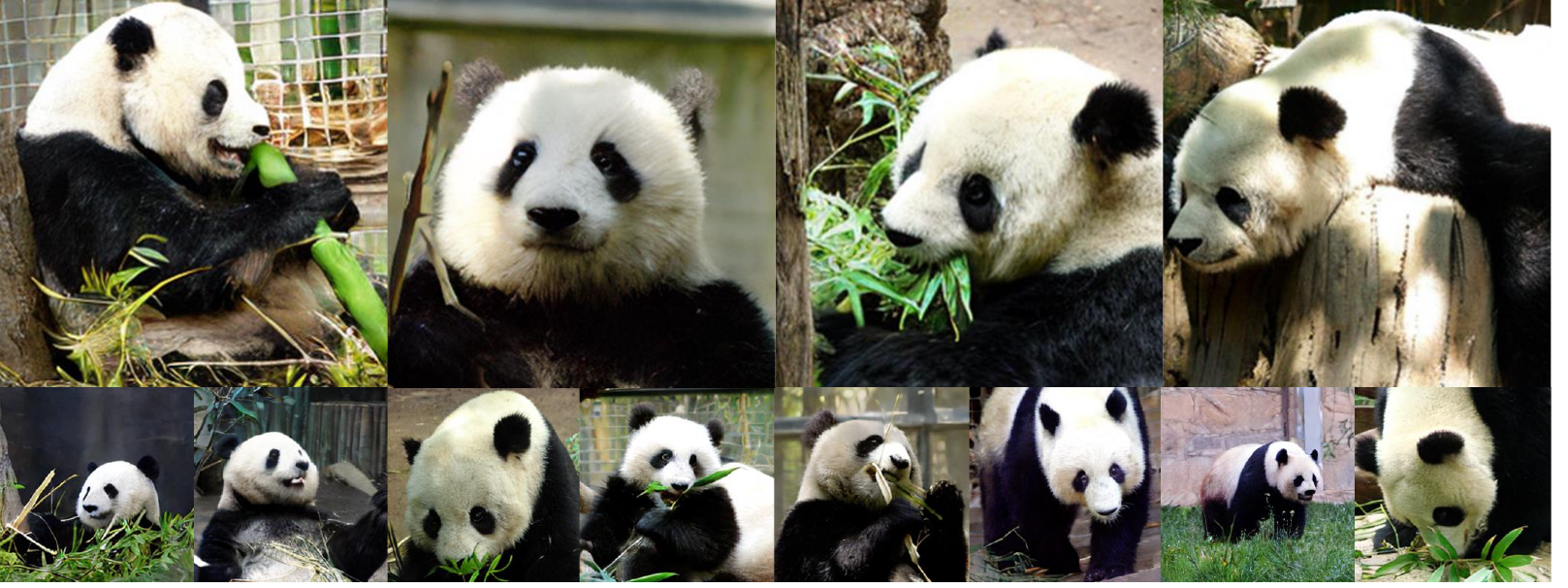}
  \vspace{-14pt}
    \caption{Generated samples of SiT-L/2 trained with \textit{D\textsuperscript{2}C} using a 50K dataset (CFG=1.5). Class label = "panda"(388)}
    
  \vspace{-5pt}
\end{figure*}

\begin{figure*}[t]
  \centering
  \includegraphics[width=1.0\linewidth]{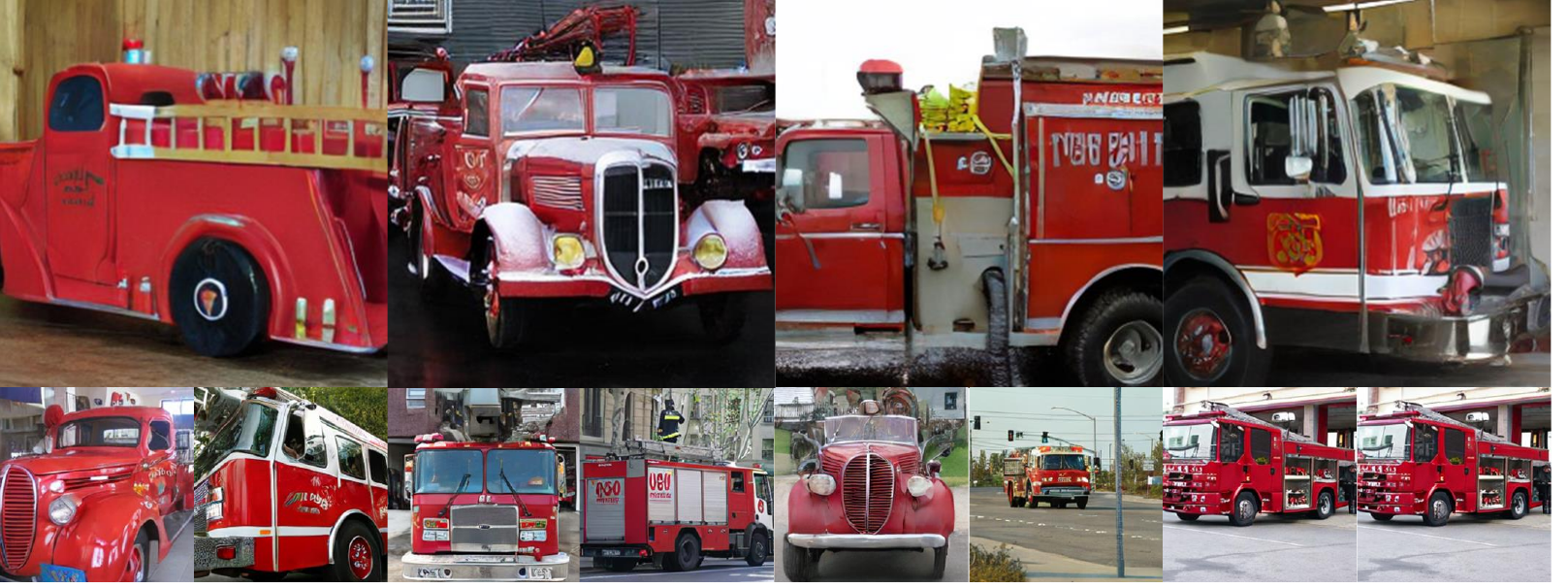}
  \vspace{-14pt}
    \caption{Generated samples of SiT-L/2 trained with \textit{D\textsuperscript{2}C} using a 50K dataset (CFG=1.5). Class label = "fire truck"(555)}
    
  \vspace{-5pt}
\end{figure*}

\begin{figure*}[t]
  \centering
  \includegraphics[width=1.0\linewidth]{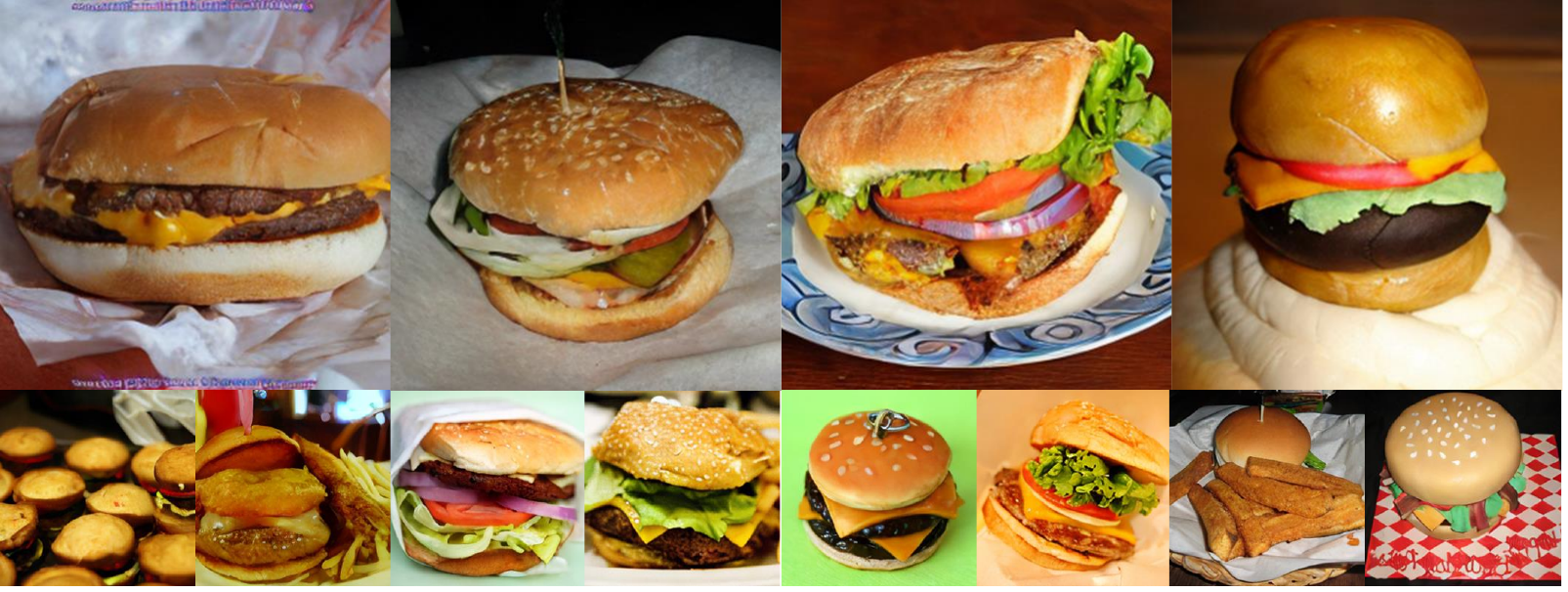}
  \vspace{-14pt}
    \caption{Generated samples of SiT-L/2 trained with \textit{D\textsuperscript{2}C} using a 50K dataset (CFG=1.5). Class label = "cheeseburger"(933)}
    
  \vspace{-5pt}
\end{figure*}

\begin{figure*}[t]
  \centering
  \includegraphics[width=1.0\linewidth]{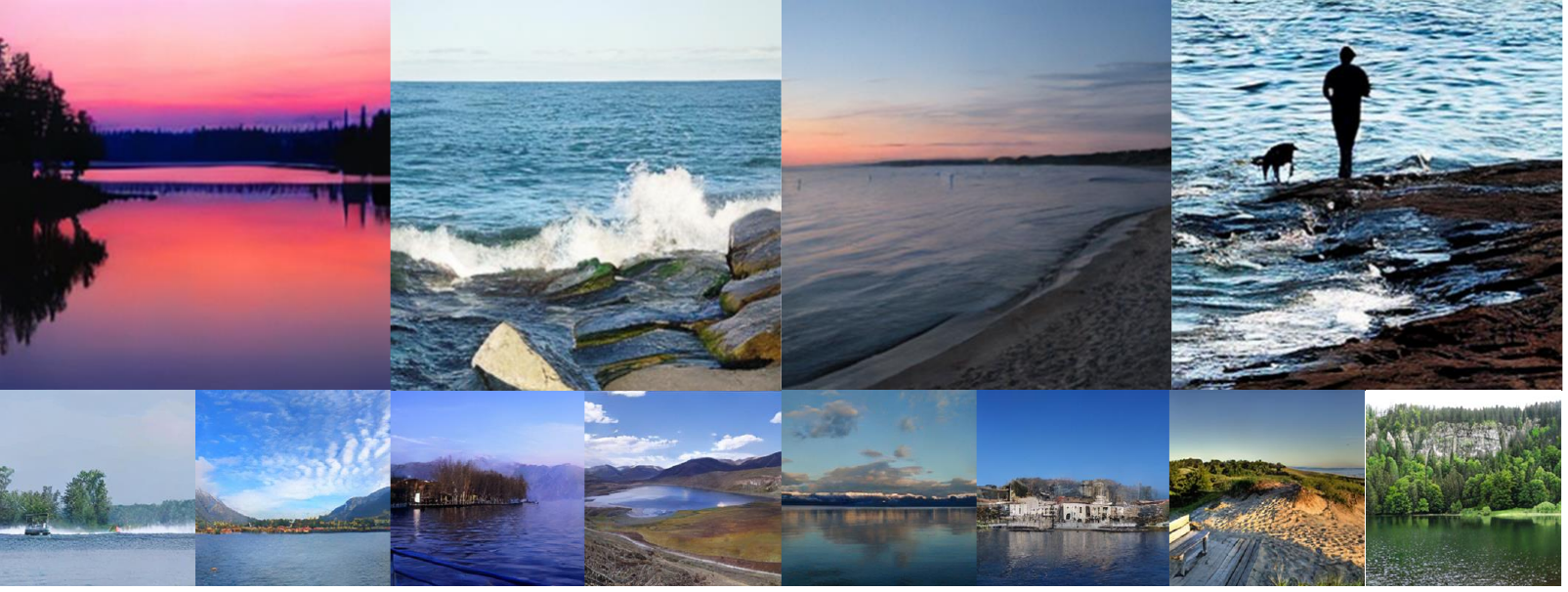}
  \vspace{-14pt}
    \caption{Generated samples of SiT-L/2 trained with \textit{D\textsuperscript{2}C} using a 50K dataset (CFG=1.5). Class label = "lake shore"(975)}
    
  \vspace{-5pt}
\end{figure*}

\begin{figure*}[t]
  \centering
  \includegraphics[width=1.0\linewidth]{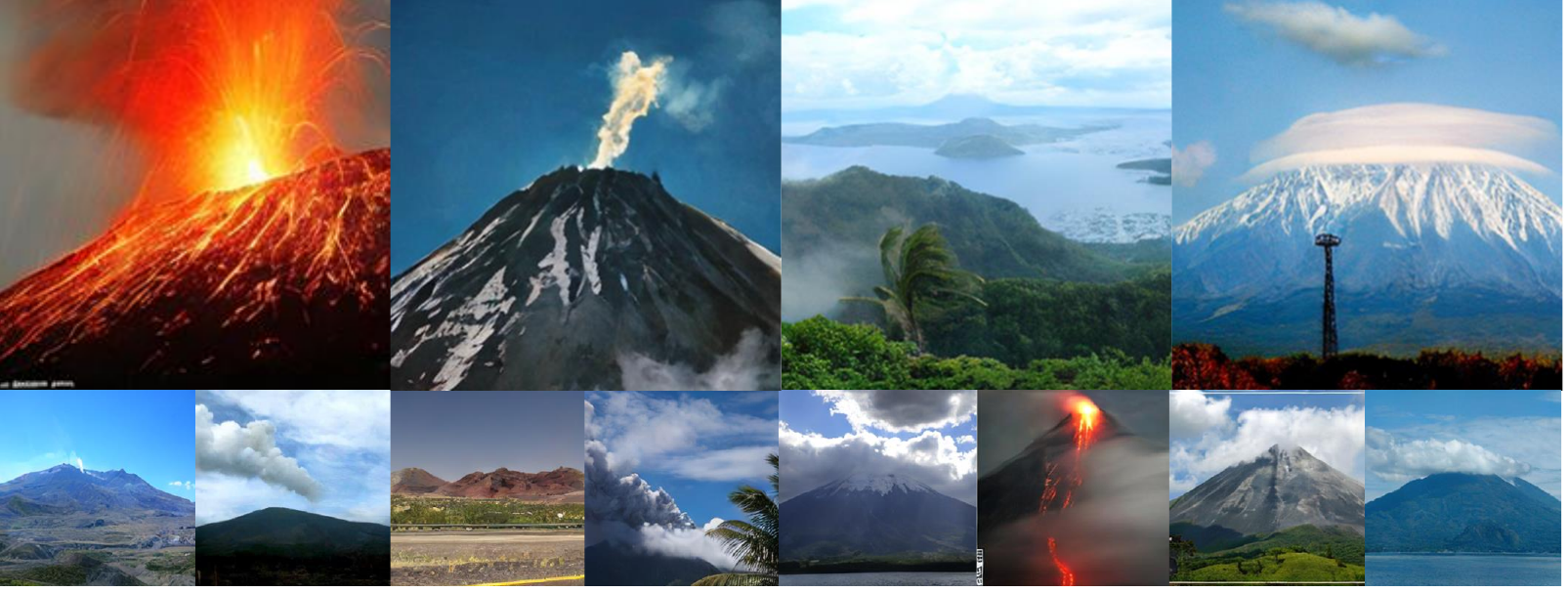}
  \vspace{-14pt}
    \caption{Generated samples of SiT-L/2 trained with \textit{D\textsuperscript{2}C} using a 50K dataset (CFG=1.5). Class label = "volcano"(980)}
    
  \vspace{-5pt}
\end{figure*}

%%%%%%%%%%%%%%%%%%%%%%%%%%%%%%%%%%%%%%%%%%%%%%%%%%%%%%%%%%%%